\documentclass[letterpaper]{article} 
\usepackage{aaai2026}  
\usepackage{times}  
\usepackage{helvet}  
\usepackage{courier}  
\usepackage[hyphens]{url}  
\usepackage{graphicx} 
\usepackage{natbib}  
\usepackage{caption} 
\usepackage{algorithm}
\usepackage{algorithmic}
\usepackage{amsmath}
\usepackage{booktabs}

\usepackage{newfloat}
\usepackage{listings}
\DeclareCaptionStyle{ruled}{labelfont=normalfont,labelsep=colon,strut=off} 
\floatstyle{ruled}
\newfloat{listing}{tb}{lst}{}
\floatname{listing}{Listing}
\title{Structure-based RNA Design by Step-wise Optimization of Latent Diffusion Model}
\author {
    Qi Si\textsuperscript{\rm 1}\equalcontrib,
    Xuyang Liu\textsuperscript{\rm 1}\equalcontrib,
    Penglei Wang\textsuperscript{\rm 2}\equalcontrib,
    Xin Guo\textsuperscript{\rm 1$\dagger$},
    Yuan Qi\textsuperscript{\rm 1,3,4},
    Yuan Cheng\textsuperscript{\rm 1,3}\thanks{Corresponding author.}
}
\affiliations {
    \textsuperscript{\rm 1}Shanghai Academy of Artificial Intelligence for Science\\
    \textsuperscript{\rm 2}School of Biomedical Engineering, Shanghai Jiao Tong University\\
    \textsuperscript{\rm 3}Artificial Intelligence Innovation and Incubation Institute, Fudan University\\
    \textsuperscript{\rm 4}Zhongshan Hospital, Fudan University\\
    sqwd0616@gmail.com, wangpenglei@sjtu.edu.cn
}

\usepackage{bibentry}

\begin{document}

\maketitle

\begin{abstract}
RNA inverse folding, designing sequences to form specific 3D structures, is critical for therapeutics, gene regulation, and synthetic biology. Current methods, focused on sequence recovery, struggle to address structural objectives like secondary structure consistency (SS), minimum free energy (MFE), and local distance difference test (LDDT), leading to suboptimal structural accuracy. To tackle this, we propose a reinforcement learning (RL) framework integrated with a latent diffusion model (LDM). Drawing inspiration from the success of diffusion models in RNA inverse folding, which adeptly model complex sequence-structure interactions, we develop an LDM incorporating pre-trained RNA-FM embeddings from a large-scale RNA model. These embeddings capture co-evolutionary patterns, markedly improving sequence recovery accuracy. However, existing approaches, including diffusion-based methods, cannot effectively handle non-differentiable structural objectives. By contrast, RL excels in this task by using policy-driven reward optimization to navigate complex, non-gradient-based objectives, offering a significant advantage over traditional methods. In summary, we propose the Step-wise Optimization of Latent Diffusion Model (SOLD), a novel RL framework that optimizes single-step noise without sampling the full diffusion trajectory, achieving efficient refinement of multiple structural objectives. Experimental results demonstrate SOLD surpasses its LDM baseline and state-of-the-art methods across all metrics, establishing a robust framework for RNA inverse folding with profound implications for biotechnological and therapeutic applications.
\end{abstract}

\begin{links}
    \link{Code}{https://github.com/darkflash03/SOLD}
\end{links}

\section{Introduction}

The RNA inverse folding task involves designing RNA sequences that fold into specific 3D structures, holds significant potential for applications in RNA therapeutics, gene regulation, and synthetic 
biology~\cite{damase2021rna}. For instance, rationally designed riboswitches can enable precise control of mRNA translation, paving the way for targeted therapies~\cite{mustafina2019design}.

Existing RNA inverse folding methods primarily include physics-based and deep learning-based methods. Physics-based approaches like Rosetta~\cite{leman2020macromolecular}. generate sequences via Monte Carlo optimization but are computationally expensive and struggle with polymorphic conformations Algorithms based on 2D structures, such as ViennaRNA~\cite{churkin2018viennarna}, achieve computational efficiency but neglect critical 3D geometric information, which restricts their overall applicability. In recent years, deep learning-based methods have gained traction, with generative models showing promise. Unlike variational autoencoders, as RhoDesign~\cite{wong2024deep} and RDesign~\cite{tan2024rdesign}, which struggle with complex distributions and long-range dependencies, diffusion models excel at capturing intricate sequence-structure interactions through iterative denoising. RiboDiffusion~\cite{huang2024ribodiffusion} proposed a diffusion-based framework for RNA inverse folding, but it depends on direct sequence modeling, which may not fully capture the rich co-evolutionary context inherent in RNA sequences. 

RL has recently been widely applied to optimize the generation objectives of diffusion models. For instance, DDPO~\cite{black2023training} and DPOK~\cite{Fan2023DPOK} combined diffusion models with policy optimization to optimize decision-making in continuous action spaces. These studies demonstrate that RL can guide diffusion models to optimize objectives that are difficult to address directly with traditional methods by designing specific reward functions. Therefore, RL is particularly well-suited for diffusion models in RNA inverse folding, as it effectively optimizes non-differentiable structural metrics. Despite its potential, RL remains underexplored in this area.

To address the above limitations, we propose SOLD, a LDM based framework for RNA inverse folding. SOLD leverages the embedding space of the RNA-FM~\cite{chen2022interpretable} to capture co-evolutionary information. Furthermore, SOLD introduces a novel step-wise RL optimization strategy that directly optimizes structural metrics (SS, MFE, LDDT), where step-wise refers to starting from any noise time step and independently optimizing each reverse sampling step. This approach ensures both computational efficiency and superior performance in generating RNA sequences that align with target structures. Our main contributions are as follows:

\begin{itemize}
    \item  A LDM that leverages pre-trained LLM embedding to improve sequence recovery in RNA design.
    \item The first integration of RL into LDM for RNA inverse folding, enabling effective optimization of complex structural objectives.
    \item  A step-wise RL optimization algorithm that achieves faster convergence and superior performance.
\end{itemize}

\section{Related Work}
\subsection{RNA Inverse Folding Methods}

RNA inverse folding designs sequences for specified structures, categorized into physics-based, heuristic, and deep learning approaches. Physics-based Rosetta ~\cite{das2010rosetta,leman2020macromolecular} uses Monte Carlo sampling with energy functions but is computationally intensive. Early 2D structure-based approaches such as RNAinverse ~\cite{hofacker1994} and ViennaRNA ~\cite{lorenz2011} were dynamically programmed directly on Rfam ~\cite{griffiths2003rfam}, thus ignoring 3D structural information. Among the heuristics, INFO-RNA~\cite{busch2006} performs a direct random search of the structure, whereas methods such as RNASSD~\cite{andronescu2004}, antaRNA~\cite{kleinkauf2015}, aRNAque~\cite{merleau2022}, eM2dRNAs~\cite{rubio-largo2023}, and MCTS-RNA~\cite{yang2017} use global search or evolutionary approaches.

Deep learning methods such as gRNAde~\cite{joshi2025grnade} is a concurrent graph-based RNA refolding method, while RhoDesign~\cite{wong2024deep}, RDesign~\cite{tan2024rdesign}, PiFold~\cite{gao2023pifold}, StructGNN~\cite{chou2024}, and GVP-GNN~\cite{jing2021learning} are representative deep learning methods for inverse folding, which are modified here for RNA compatibility. Recently,  RISoTTo~\cite{bibekar2025context} proposed a context-aware geometric deep learning framework that integrates structural context into RNA sequence design. While these deep learning approaches enhance the quality of generated sequences, they are unable to directly optimize structure metrics, such as MFE and LDDT.

\subsection{Diffusion Models in Molecular Generation}
 RiboDiffusion~\cite{huang2024ribodiffusion} applies 3D diffusion on PDB dataset for RNA inverse folding. DRAKES~\cite{wang2024finetuning} fine-tunes discrete diffusion with reinforcement learning, optimizing DNA enhancer activity and protein stability. GradeIF~\cite{yi2023graph} leverages graph denoising with BLOSUM matrices, enhancing protein sequence diversity and recovery. RNAdiffusion~\cite{huang2024rnadiffusion} uses latent diffusion to generate RNA sequences, optimizing the translation efficiency. Structured DDPM~\cite{austin2021} and Dirichlet Flow~\cite{stark2024} explored discrete diffusion on molecular data. RNAFlow~\cite{chen2024rnaflow} integrates 2D/3D diffusion in RNA structures. These models address diverse molecular design challenges, from structure prediction to functional optimization.

\subsection{Reinforcement Learning for Generative Model Optimization}

RL has emerged as a powerful approach for optimizing generative models, particularly diffusion models, by aligning them with diverse objectives. Online RL methods like PPO~\cite{schulman2017proximal} and RAFT~\cite{dong2023raft} leverage direct reward optimization, while offline methods such as DPO~\cite{rafailov2023direct} utilize preference datasets for policy alignment. 

Meanwhile, alignment of difffusion models to preferences by reinforcement learning has also been widely explored. DDPO~\cite{black2023training} and DPOK~\cite{Fan2023DPOK} use the PPO algorithm to enhance image quality in diffusion models; Diffusion-DPO~\cite{wallace2023diffusion} adapts DPO to fine-tune Stable Diffusion XL with human preferences. These studies demonstrate the ability of RL to guide models toward optimizing objectives that traditional methods struggle to address directly.

\subsection{Comparison with Our Work}

SOLD integrates latent diffusion and RL to advance RNA inverse folding, offering some significant advantages over existing methods:

\begin{itemize}
    \item Unlike RNAdiffusion, which generates sequences without structural constraints and optimizes functional metrics using separately trained reward models, SOLD focuses on structure-based RNA design, directly optimizing structural metrics. By leveraging ViennaRNA for direct evaluation, SOLD eliminates the need for additional reward models, reducing computational overhead and mitigating inaccuracies from reward model predictions.
    
    \item DRAKES utilizes discrete diffusion with RL to optimize a single objective for RNA design, relying on differentiable reward models that introduce additional training costs and the potential risk of over-optimization. However, SOLD operates in a continuous latent space for diffusion, enabling efficient optimization of multiple RNA-specific metrics without the need for separately trained reward models.

    \item In contrast to DDPO and DPOK, which are designed for image and text generation and rely on sampling complete trajectories, incurring high computational costs, SOLD is tailored for RNA inverse folding, and its step-wise RL optimization iteratively refines strategies, achieving greater efficiency compared to trajectory-based methods.
\end{itemize}

\section{Methods}

\begin{figure*}[t]
    \vspace*{-10pt} 
    \centering
    \includegraphics[width=\textwidth]{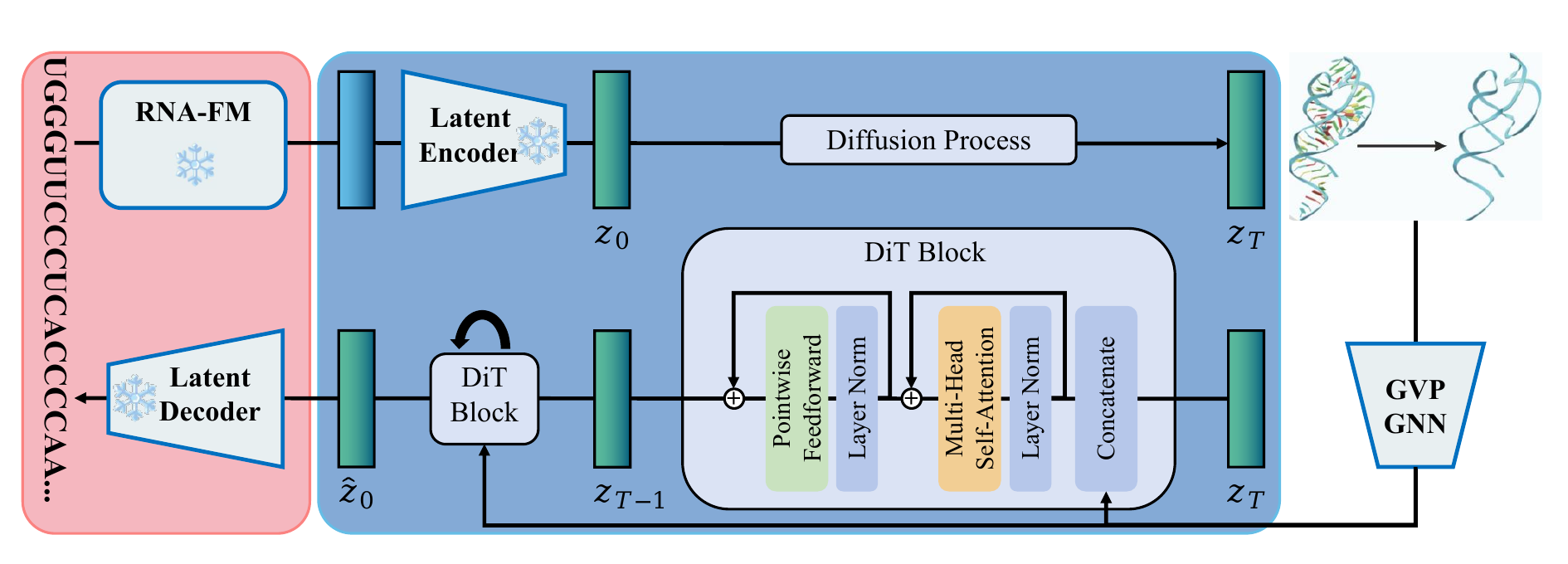}
    \caption{The LDM encodes RNA-FM embeddings into a latent representation, performs denoising with GVP-GNN + DiT blocks, and decodes the refined latent into RNA sequences consistent with structural constraints.}
    \label{fig:LDM}
\end{figure*}

\subsection{Latent Diffusion Model}

The SOLD employs a latent diffusion model (LDM) to generate RNA sequences, integrating pre-trained RNA-FM to extract embeddings with geometric information from the backbone, producing sequences that conform to specific structural and functional properties as Figure \ref{fig:LDM}. We map RNA sequences directly into variable-length embeddings of shape $(L, 640)$, where $L$ represents the sequence length. These embeddings are compressed into a latent representation of shape $(L, D)$ via an MLP encoder to improve the efficiency of the diffusion process. 

A denoising network $\pi_\theta$, integrated with GVP-GNN~\cite{jing2021learning} and Diffusion Transformer (DiT)~~\cite{peebles2023scalable}, is used to predict the denoised latent embedding $\hat{z}_0$ at time $t=0$. The MLP decoder reconstructs the latent embedding $\hat{z}_0$ into a sequence probability distribution $(L, 4)$, generating sequences of length $L$ that align with the input backbone structure.

The forward process of diffusion introduces Gaussian noise to the latent embedding $z_0 \in \mathbf{R}^{L \times D}$:
\begin{equation}
q(z_t | z_{t-1}) = \mathcal{N}(z_t; \sqrt{1-\beta_t} z_{t-1}, \beta_t I)
\label{eq:forward_diffusion}
\end{equation}
where $\beta_t$ is the noise schedule, and $t=1,\dots,T$. The reverse process reconstructs the embedding by predicting $\hat{z}_0 = \pi_\theta(z_t, t, c)$:
\begin{equation}
p_\theta(z_{t-1} | z_t, c) = \mathcal{N}(z_{t-1}; \mu_\theta(z_t, t, c), \sigma_t^2 I)
\label{eq:reverse_diffusion}
\end{equation}
with the mean defined as:
\begin{equation}
\mu_\theta(z_t, t, c) = \frac{\sqrt{\alpha_t} (1-\bar{\alpha}_{t-1})}{1-\bar{\alpha}_t} z_t + \frac{\sqrt{\bar{\alpha}_{t-1}} (1-\alpha_t)}{1-\bar{\alpha}_t} \hat{z}_0
\label{eq:reverse_mean}
\end{equation}
where $c$ represents the geometric information from backbone, $\alpha_t = 1-\beta_t$, $\bar{\alpha}_t = \prod_{i=1}^t \alpha_i$, and the variance is defined as $\sigma_t^2 = \frac{(1-\bar{\alpha}_{t-1})(1-\alpha_t)}{1-\bar{\alpha}_t}$.

The training objective combines MSE and cross-entropy:
\begin{equation}
\mathcal{L} =
\mathrm{E}_{z_0,t}\!\bigl[\|z_0-\hat z_0\|^2\bigr]
- \mathrm{E}_{s}\!\left[\sum_{i=1}^{L} \log p\!\left(s_i \mid \mathrm{Dec}(\hat z_0)_i\right)\right]
\label{eq:loss}
\end{equation}

The first term of the loss function, denoted as the mean squared error (MSE) term, quantifies the difference between the true latent embedding $z_0$ and the single-step predicted embedding $\hat{z}_0$, enhancing the precision of  denoise network in predicting the clean embedding. The second term referred to the cross-entropy term, ensures that the sequence decoded from the final denoised embedding $\hat{z}_0$ closely aligns with the target sequence $s$, guiding the frozen decoder to generate accurate nucleotide probability distributions.

The encoder efficiently compresses RNA-FM embeddings into the latent space, thereby facilitating the diffusion process.
\begin{equation}
z_0 = \mathrm{Enc}_\psi(h) = \mathrm{MLP}_\psi(h)
\label{eq:encoder}
\end{equation}
where $h \in \mathbf{R}^{L \times 640}$ represents the input embedding, and $z_0 \in \mathbf{R}^{L \times D}$ is the compressed latent representation. The decoder generates sequence probabilities by:
\begin{equation}
p(s) = \mathrm{Dec}_\phi(\hat{z_0}) = \mathrm{softmax}(\mathrm{MLP}_\phi(\hat{z_0}))
\label{eq:decoder}
\end{equation}
where $\hat{z_0} \in \mathbf{R}^{L \times D}$ is the final denoised latent embedding, and $p(s) \in \mathbf{R}^{L \times 4}$ represents the probability distribution over the four nucleotide bases (A, U, C, G).

\begin{figure*}[t]
    \vspace*{-10pt} 
    \centering
    \small
    \includegraphics[width=\textwidth]{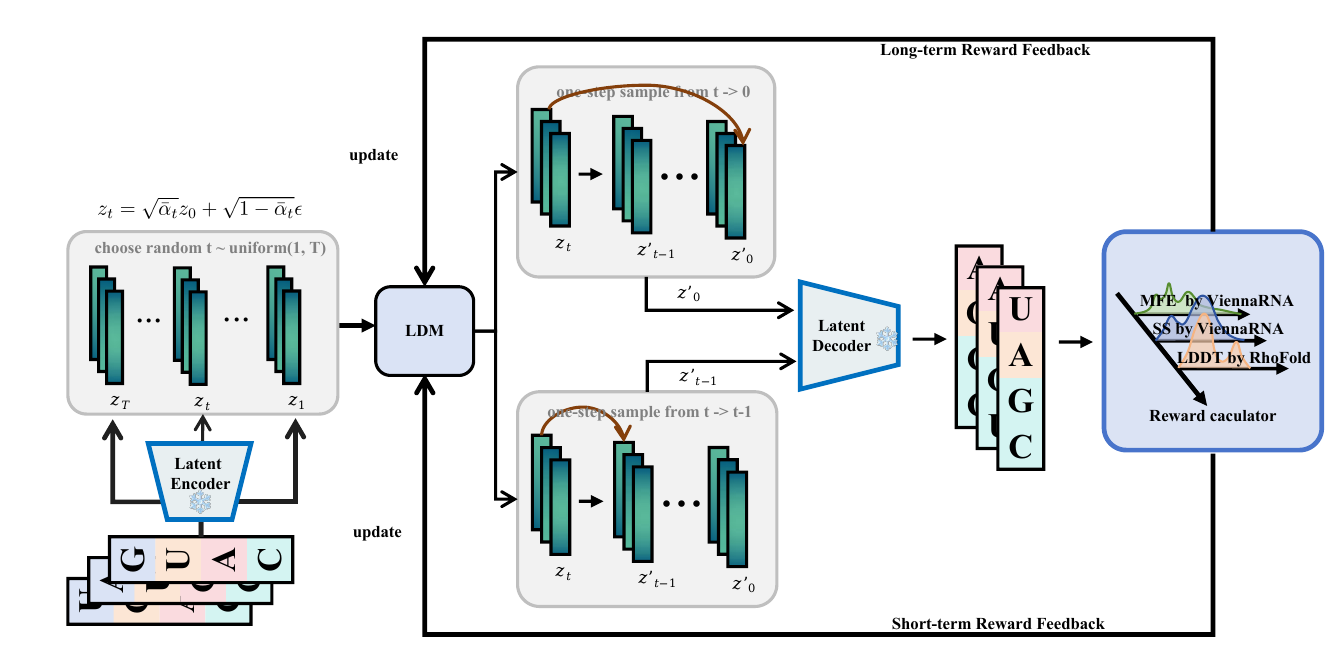}
    \caption{An overview of the SOLD. SOLD utilizes long-term and short-term reward feedback to directly optimize trained latent diffusion model in a random denoising step.}
    \label{fig:schema}
\end{figure*}

\subsection{Step-wise Optimization of Latent Diffusion Model}

The SOLD algorithm refines the pre-trained LDM through a RL framework, targeting complex objectives critical for RNA inverse folding, such as SS, MFE, and LDDT. The overall architecture is depicted in Figure~\ref{fig:schema}. SOLD adopts a single-step sampling strategy inspired by Denoising Diffusion Implicit Models (DDIM)~\cite{song2020denoising} to enhance computational efficiency. In DDIM, the reverse process is approximated with a deterministic mapping, enabling single-step predictions of the target latent embedding \( \hat{z}_0 \) from a noisy sample \( z_t \). 

SOLD's single-step prediction leverages the Denoising Diffusion Implicit Model (DDIM) framework to efficiently generate denoised latent embeddings. The sampling process is defined as:
\begin{equation}
z_{t-k}' = \sqrt{\bar{\alpha}_{t-k}}\,\hat{z}_0
  + \gamma_{t-k}\,\epsilon_\theta(z_t, t, c)
  + \sigma_{t-k}\,\epsilon
\label{eq:ddim}
\end{equation}
where \(\hat{z}_0 = \pi_\theta(z_t, t, c)\) is the predicted target embedding at \(t=0\), $\gamma_{t-k} = \sqrt{1 - \bar{\alpha}_{t-k} - \sigma_{t-k}^2}$ is a shorthand coefficient, \(
\epsilon_\theta(z_t, t, c) = \frac{z_t - \hat{z}_0 \sqrt{\bar{\alpha}_t}}{\sqrt{1 - \bar{\alpha}_t}}
\) is the predicted noise conditioned on structural constraints \(c\), and \(\epsilon \sim \mathcal{N}(0, I)\) is standard Gaussian noise. The stochasticity is controlled by \(\sigma_{t-k}^2 = \eta \cdot \frac{(1 - \bar{\alpha}_{t-k})(1 - \alpha_t)}{1 - \bar{\alpha}_t}\), with hyperparameter \(\eta\). When \(k=t\), SOLD can sample \(z_0'\) in a single step, significantly enhancing computational efficiency. By decoding \(z_0'\) into a sequence \(s_0'\), SOLD evaluates a long-term reward \(r_0(t) = R_i(s_0')\), where \(R_i\) assesses RNA-specific objectives.

By randomly sampling a timestep \( t \) and directly predicting \( z_0' \) from \( z_t \), SOLD eliminates the need to generate complete trajectories during training, reducing computational complexity. Notably, while SOLD uses single-step sampling for training efficiency, it retains full trajectory denoising during inference to ensure high-quality sequence generation.

In early denoising steps (large \( t \)), the smaller \(\bar{\alpha}_t\) increases the variance of the noise term, reducing the accuracy of direct \( z_0' \) predictions and the reliability of long-term rewards \( r_0(t) \). To address this, SOLD computes short-term rewards by equation~\ref{eq:ddim} with sampling strategy (\( k=1 \)) to predict the intermediate latent embedding \( z_{t-1}' \) from \( z_t \). By decoding \( z_{t-1}' \) into a sequence \( s_{t-1}' \), SOLD computes a short-term reward \( r_t(t) = R_i(s_{t-1}') \), using the same reward function \( R_i \) as the long-term reward. This short-term rewards effectively guide learning during early denoising steps where high noise levels diminish the reliability of long-term rewards.

To balance the strengths of both rewards, SOLD integrates long-term and short-term rewards through a piecewise reward function:
\begin{equation}
r_{\mathrm{total}}(t) = w(t) \cdot r_t(t) + u(t) \cdot r_0(t)
\label{eq:rewards}
\end{equation}
where \( w(t) \) and \( u(t) \) are time-dependent weighting functions. In early denoising steps (large \( t \)), \( w(t)=1 \) and \( u(t)=0 \), prioritizing the short-term reward \( r_t(t) \) to guide learning. In later denoising steps (small \( t \)), \( w(t)=0 \) and \( u(t)=1 \), emphasizing the long-term reward \( r_0(t) \). This piecewise strategy enhances SOLD’s flexibility in optimizing RNA-specific objectives across different denoising stages.

\begin{table*}[t]
\centering
\begin{tabular}{lcccc}
\hline
\multicolumn{1}{c}{} & \multicolumn{2}{c}{SOLD TEST} & \multicolumn{2}{c}{CASP15 TEST} \\
\hline
Method         & Sequence Recovery & NT Recovery & Sequence Recovery & NT Recovery \\
\hline
RhoDesign      & 0.2734       & 0.2859      & 0.2606       & 0.2575      \\
RDesign        & 0.4457       & 0.3966      & 0.3264       & 0.3251      \\
gRNAde         & 0.5108       & 0.4890      & 0.5097       & 0.5149      \\
RiboDiffusion  & 0.5125       & 0.4416      & 0.5388       & 0.3871      \\
DRAKES-Pretrain & 0.4524        & 0.4088      & 0.3357       & 0.3374      \\
LDM            & \textbf{0.5728}  & \textbf{0.5034} & \textbf{0.5462}  & \textbf{0.5473} \\
\hline
\end{tabular}
\caption{Sequence Recovery and NT Recovery Comparison}
\label{tab:seq_aa_recovery}
\end{table*}

SOLD optimizes the model parameters \( \theta \) using Proximal Policy Optimization (PPO)~\cite{schulman2017proximal} based on the total reward \( r_{\mathrm{total}}(t) \), with the objective:
\begin{equation}
\mathcal{J}_{\mathrm{SOLD}}(\theta) = \mathrm{E}_{t \sim \mathcal{U}[1, T], c, z_t, z_0' \sim p_\theta(z_0' \mid z_t, c)} \left[ r_{\mathrm{total}}(t) \right]
\label{eq:sold_objective}
\end{equation}
where the policy gradient is computed via the REINFORCE algorithm with importance sampling:
\begin{equation}
\nabla_\theta \mathcal{J}_{\mathrm{SOLD}} = \mathrm{E}_{p_{\mathrm{old}}(t)} \left[ w \nabla_\theta \log p_\theta(z_0' \mid z_t, t, c) r(z_t) \right]
\label{eq:sold_gradient}
\end{equation}

where the importance weight is:
\begin{equation}
w = \frac{p_\theta(x_0' \mid x_t, t, c)}{p_{\mathrm{ref}}(x_0' \mid x_t, t, c)}
\end{equation}
with \( p_{\mathrm{ref}}(z_0' \mid z_t, t, c) \) representing the conditional probability, under the previous iteration’s model parameters \( \theta_{\mathrm{ref}} \). 

SOLD adopts a clipped surrogate objective, with a clip range of \(\epsilon = 0.0001\), to ensure stable policy updates. Additionally, SOLD incorporates a constraint to limit divergence between the current policy \( p_\theta \) and the previous policy \(p_{\mathrm{ref}}\):
\begin{equation}
\mathcal{J}_{\mathrm{ref}}(\theta) = -D_{\mathrm{KL}}(p_\theta \| p_{\mathrm{ref}})
\label{eq:kl_ref}
\end{equation}
Finally, the total optimization objective is:
\begin{equation}
\mathcal{J}_{\mathrm{SOLD}}^{\mathrm{reg}}(\theta) = \mathcal{J}_{\mathrm{SOLD}}(\theta) + \lambda_{\mathrm{ref}} \mathcal{J}_{\mathrm{ref}}(\theta)
\label{eq:kl_regularization}
\end{equation}
where \(\lambda_{\mathrm{ref}}\) is the regularization weight for the reference policy. This integration of single-step sampling, piecewise reward framework, and KL-constrained optimization enables SOLD to effectively tackle complex RNA inverse folding design.

\section{Experiment}

To evaluate the performance of SOLD for RNA inverse folding, we conducted a series of experiments. Our experiments leverage a high-quality dataset constructed from the RCSB Protein Data Bank, RNAsolo, and CASP15 RNA dataset, as detailed in Appendix A. The dataset was pre-processed by clustering RNA structural data using PSI-CD-HIT (sequence threshold: 0.3) and US-align (structural threshold: 0.45). After pre-processing, we obtained 8222 structures, split into pre-training (7067 structures), RL fine-tuning (389 structures), and SOLD TEST (766 structures) datasets, with deduplication to prevent information leakage. The CASP15 TEST dataset served as an independent test benchmark.

We compared SOLD against state-of-the-art (SOTA) methods, including RhoDesign~\cite{wong2024deep}, RDesign~\cite{tan2024rdesign}, gRNAde~\cite{joshi2025grnade}, RiboDiffusion~\cite{huang2024ribodiffusion}, and DRAKES~\cite{wang2024finetuning}. Notably, DRAKES-Pretrain refers to the pre-trained DRAKES model without RL finetuning, while DRAKES incorporates optimization for a single metric (MFE), using a discrete diffusion framework with differentiable reward models. In contrast, SOLD employs LDM with step-wise RL optimization, targeting multiple structurally relevant metrics without any differentiable reward model.

The experiments are structured in three parts: (1) evaluating the LDM’s superiority in 1D metrics (Sequence Recovery, Nucleotide (NT) Recovery) to test its foundational generative capability; (2) assessing SOLD’s RL finetuning for single-objective optimization, evaluating its speed and effectiveness in optimizing 2D (SS, MFE) and 3D (LDDT) metrics; and (3) validating the effectiveness of SOLD’s weighted multi-objective optimization across 1D (Sequence Recovery), 2D (SS, MFE), and 3D (RMSD, LDDT) metrics, with a practical case study to examine SOLD’s real-world utility in RNA design. All experiments were conducted on a single A100 GPU.

\subsection{LDM Performance on Sequence Generation}

To assess the effectiveness of the LDM backbone in SOLD, we evaluated its sequence generation performance using 1D metrics, specifically Sequence Recovery and NT Recovery, which measure the similarity of generated sequences to target sequences and the accuracy of nucleotide composition, respectively. These metrics focus on the model’s ability to predict the correct nucleotide at each position, with structural information serving as input features that implicitly influence sequence generation rather than as explicit optimization targets. As the pre-training phase of LDM focuses on optimizing sequence prediction rather than directly targeting 2D or 3D structural metrics, we evaluate performance only using 1D metrics. Table~\ref{tab:seq_aa_recovery} presents the results on the SOLD TEST and CASP15 TEST datasets, comparing SOLD’s LDM against RhoDesign, RDesign, gRNAde, RiboDiffusion, and DRAKES-Pretrain. Since this evaluation precedes RL finetuning, we compare against DRAKES-Pretrain to ensure a fair assessment of pre-trained models. LDM consistently outperforms competitors, showing higher Sequence Recovery and NT Recovery across both test datasets. Compared with RiboDiffusion, which performs SDE-based diffusion directly in the one-hot RNA sequence space, our LDM operates in the RNA-FM embedding space while keeping the backbone architecture unchanged. This latent representation captures richer structural and co-evolutionary features, leading to markedly improved sequence recovery. This method appears to enhance sequence prediction accuracy while maintaining computational efficiency, establishing a strong foundation for subsequent RL finetuning.

\subsection{RL Finetuning for Single-Objective Optimization}

We evaluated SOLD’s RL finetuning stage for single-objective optimization, targeting on 2D (SS, MFE) and 3D (LDDT) metrics, which are critical for RNA functionality but challenging to optimize directly in SOTA RNA design algorithms. Table~\ref{tab:single_objective} presents the results of single-objective optimization for MFE, SS, and LDDT on the SOLD TEST and CASP15 TEST datasets, comparing SOLD against its LDM baseline, DDPO, and DPOK. For MFE optimization, we additionally compared the DRAKES-Pretrain and DRAKES methods to elucidate the impact of reinforcement learning. As DRAKES exclusively optimizes MFE, it was not included in comparisons for SS and LDDT.

\begin{table}[h]
\centering
\begin{tabular}{lcc}
\hline
Method         & SOLD TEST & CASP15 TEST \\
\hline
\multicolumn{3}{c}{\textbf{MFE Training (MFE reward) \(\downarrow\)}} \\
\hline
DRAKES-Pretrain& -12.5123               & -52.6827                   \\

DRAKES         & -14.2374          & -61.0354            \\

LDM            & -13.1519          & -52.7387           \\

DDPO           & -18.7498                  & -63.9567                 \\

DPOK           & -17.4660                  & -67.7949           \\

SOLD           & \textbf{-19.7428}     & \textbf{-68.2100}           \\
\hline
\multicolumn{3}{c}{\textbf{SS Training (SS reward) \(\uparrow\)}} \\
\hline
LDM            & 0.7274          & 0.5543           \\

DDPO           & \textbf{0.7595}                 & 0.6649                \\

DPOK           & 0.7511                 & 0.6303           \\

SOLD           & 0.7551    & \textbf{0.7010}           \\
\hline
\multicolumn{3}{c}{\textbf{LDDT Training (LDDT reward) \(\uparrow\) }} \\
\hline
LDM            & 0.6184          & 0.3237           \\

DDPO           & 0.6286          & 0.3406           \\

DPOK           & 0.6329           & 0.3351           \\

SOLD           & \textbf{0.6384}          & \textbf{0.3548}           \\
\hline
\end{tabular}
\caption{Single-Objective Reward Comparison}
\label{tab:single_objective}
\end{table}

\begin{figure*}[t]
    \vspace*{-10pt} 
    \centering
    \includegraphics[width=\textwidth, height=0.6\textheight, keepaspectratio]{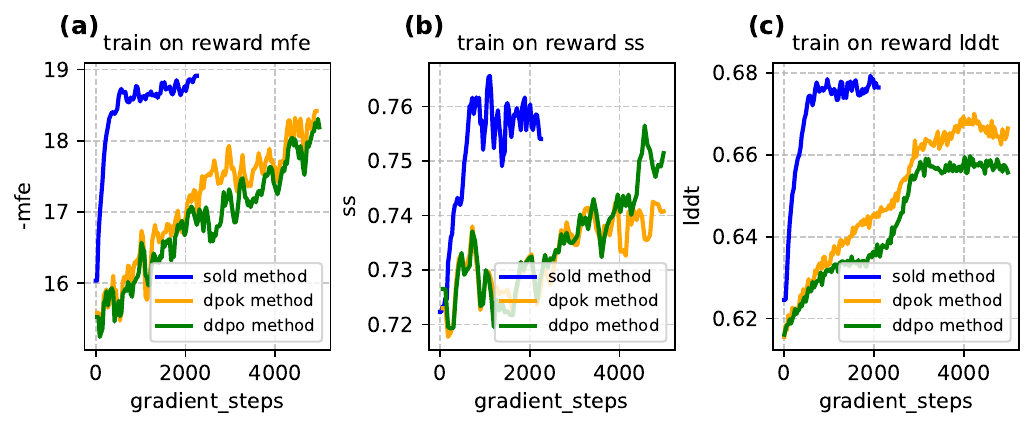}
    \caption{rewards for SOLD, DPOK, and DDPO with MFE, SS, and LDDT as reward objectives: (a) MFE, (b) SS, (c) LDDT.}
    \label{fig:methods_ablation}
\end{figure*}

For MFE, SOLD demonstrates a better improvement over its LDM baseline compared to the improvement of DRAKES over DRAKES-Pretrain, while also achieving superior final performance. Across all objectives (MFE, SS, LDDT), SOLD consistently improves upon LDM, with varying degrees of enhancement depending on the metric. Figure~\ref{fig:methods_ablation} illustrates the rapid convergence of SOLD compared to DDPO and DPOK, highlighting its efficiency in optimizing these structurally relevant metrics. Table~\ref{tab:methods_training_time} further quantifies this efficiency, revealing that SOLD completes a single training round for each objective much faster than DDPO and DPOK. For LDDT, SOLD’s training speed is slightly slower due to the computational bottleneck of structure prediction, but it still remains competitive. Table~\ref{tab:single_objective} further confirms that the performance of SOLD on MFE, SS, and LDDT is generally on par with or superior to DDPO and DPOK, demonstrating its effectiveness alongside its speed advantage. 

\begin{table}
    \centering
    \begin{tabular}{lccc}
        \hline
        Method & MFE (s) & SS (s) & LDDT (s) \\
        \hline
        DDPO      & 5953 & 6190 & 14000 \\
        DPOK      & 7677 & 7330 & 14200 \\
        SOLD      & \textbf{256} & \textbf{263} & \textbf{6900} \\
        \hline
    \end{tabular}
    \caption{Average training time per epoch for DDPO, DPOK, and SOLD across different objectives}
    \label{tab:methods_training_time}
\end{table}

SOLD’s step-wise RL optimization, utilizing single-step sampling, accelerates convergence compared to trajectory-based methods like DDPO and DPOK. Additionally, the integration of ViennaRNA for direct reward evaluation eliminates the need for separate reward models, enhancing both efficiency and effectiveness in optimizing these challenging metrics.

\subsection{Multi-Objective Optimization Performance}

To validate SOLD's capability for weighted multi-objective optimization, we conducted experiments targeting the simultaneous improvement of 1D, 2D, and 3D metrics. To ensure generality, we applied this weighting without tuning hyperparameters. We optimize SS, MFE, and LDDT using an equal weighting scheme, addressing the differing scales by mapping MFE to the (0,1) interval with the function:
\begin{equation}
\mathrm{reward}_{\mathrm{MFE}} = \exp\left(\frac{1}{\mathrm{MFE} - \frac{1}{4}}\right)
\end{equation}

\begin{table*}[h]
\centering
\begin{tabular}{lccccc}
\hline
\multicolumn{6}{c}{\textbf{SOLD TEST}} \\
\hline
Method         & Sequence Recovery \(\uparrow\) & MFE \(\downarrow\)    & SS \(\uparrow\)    & RMSD \(\downarrow\)  & LDDT \(\uparrow\)  \\
\hline
RhoDesign      & 0.2734       & -11.9212  & 0.6499 & 16.3577 & 0.5031 \\

RDesign        & 0.4457       & -10.6990 & 0.6135 & 16.1315 & 0.5238 \\

gRNAde         & 0.5108       & -10.5409 & 0.5624 & 17.9960 & 0.4848 \\

RiboDiffusion  & 0.5125   & -15.2128   & 0.7632  & 12.3170  & 0.6102 \\

DRAKES         & 0.4400            & -14.2374     & \textbf{0.7691}      & 11.9077     & 0.6191      \\

LDM            & 0.5728    & -13.3275     & 0.7269      & 12.5732     & 0.6178      \\

SOLD           & \textbf{0.5732}  & \textbf{-16.8611}  & 0.7601 & \textbf{11.8612} & \textbf{0.6360}  \\
\hline
\multicolumn{6}{c}{\textbf{CASP15 TEST}} \\
\hline
RhoDesign      & 0.2606       & -45.9620 & 0.4708 & 30.2765 & 0.3306 \\

RDesign        & 0.3264       & -37.0679 & 0.4239 & 30.0221  & 0.3165 \\

gRNAde         & 0.5097       & -40.8491 & 0.4186  & 30.8809 & 0.3121 \\

RiboDiffusion  & 0.5388       & -41.1593  & 0.4699  & 29.5904 & 0.3210 \\

DRAKES         & 0.3484            & -61.0354      & 0.6412      & 27.1322     & 0.3587      \\

LDM            & 0.5462            & -52.7387      & 0.5543      & 29.7038     & 0.3237      \\

SOLD           & \textbf{0.5888}  & \textbf{-64.0375} & \textbf{0.6957} & \textbf{26.8422}  & \textbf{0.3680} \\
\hline
\end{tabular}
\caption{Multi-Objective Performance Comparison}
\label{tab:multi_objective}
\end{table*}

\begin{figure*}[h]
  \centering
  \includegraphics[width=\textwidth]{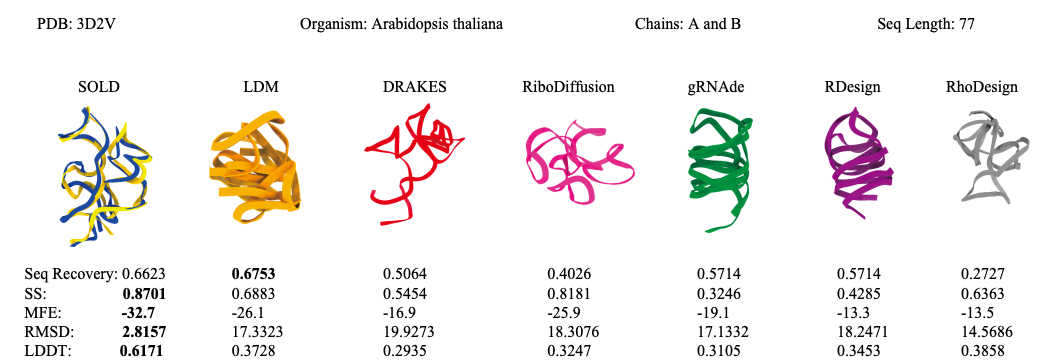}
  \caption{Comparison of rna design methods for example (PDB: 3D2V), ground-truth structure (gold), SOLD (blue) }
  \label{fig:good_case}
\end{figure*}

Table~\ref{tab:multi_objective} compares SOLD against SOTA RNA design methods on the SOLD TEST and CASP15 TEST datasets. SOLD consistently outperforms existing methods across nearly all metrics, demonstrating a superior balance between sequence naturalness and structural fidelity. Compared to its LDM baseline, SOLD maintains or improves sequence recovery with notable improvements in SS, MFE, LDDT, and RMSD. This balanced enhancement underscores SOLD's ability to optimize multiple objectives without sacrificing any single metric.
The practical efficacy of SOLD is further demonstrated with a TPP-specific riboswitch case study (Figure~\ref{fig:good_case}). Here, SOLD successfully designed a sequence that folds into the precise target structure, whereas other methods failed, producing conformations distant from the goal. This case confirms SOLD's capacity to satisfy stringent structural and functional constraints.

\section{Conclusion}
In this paper, we introduce SOLD, a novel RNA inverse folding framework that integrates a latent diffusion model (LDM) with reinforcement learning (RL) for superior performance. Combining RNA-FM embeddings with 3D backbone geometric features, SOLD's LDM excels in sequence recovery, laying a strong foundation for sequence generation. A step-wise RL algorithm further enhances performance over the LDM baseline, optimizing structurally relevant metrics with improved convergence efficiency compared to RL methods requiring full trajectory sampling. This approach represents a highly competitive algorithm that balances sequence naturalness with structural fidelity, positioning SOLD as a powerful tool for RNA design with potential for applications in therapeutics and biotechnology.

However, SOLD's performance is constrained by the limited availability of high-quality RNA structural data. Additionally, we have not extensively explored how 1D, 2D, and 3D metrics interact and coordinate in RNA design. Moreover, current reward evaluation tools, such as ViennaRNA and RhoFold, inevitably introduce approximation errors that may affect optimization accuracy. Nonetheless, our proposed framework is modular and tool-agnostic — the reward component is fully pluggable, allowing seamless substitution with more accurate structure prediction or energy evaluation models as they become available. In future work, we aim to expand datasets, refine reward evaluation with improved predictors, and explore synergistic optimization of multi-scale metrics, further enhancing SOLD's robustness and generalizability for diverse RNA design challenges.

\section{Acknowledgments}
This work was supported by the National Natural Science Foundation of China (Grant Nos. 82394432 and 92249302), 
and the Shanghai Municipal Science and Technology Major Project (Grant No. 2023SHZDZX02). 
The authors acknowledge support from the AI for Science Program, 
Shanghai Municipal Commission of Economy and Information.

\bibliography{aaai2026}

\clearpage %
\appendix
\section{A. Dataset Details}
To support the SOLD in generating RNA sequences with specific structural and functional properties for RNA inverse folding, we constructed a high-quality dataset from the RCSB Protein Data Bank~\cite{Berman2000}, RNAsolo~\cite{Skwark2022}, and CASP15~\cite{Das2023} databases. The preparation began with clustering RNA structural data from RCSB and RNAsolo on both sequence and structural aspects. Sequence clustering was performed using PSI-CD-HIT~\cite{Li2006} with a threshold of 0.3, yielding 2938 initial clusters, while structural clustering used US-align~\cite{Zhang2022} with a threshold of 0.45, producing 1462 clusters. To reserve CASP15 as an external test dataset(named \textbf{CASP15 TEST}), we removed CASP15-related cluster IDs from both sequence and structural clustering files, retaining 13,862 structures and 13,424 sequences. Merging the clustering information resulted in 13,166 structures, which were split based on structural clusters into \textbf{Pre-training}, \textbf{RL Fine-tuning}, and \textbf{SOLD TEST} dataset at a 7:1:2 ratio. Sequence clustering information was used to deduplicate the RL Fine-tuning and SOLD TEST datasets against the Pre-training dataset, and the SOLD TEST dataset was further deduplicated using the RL Fine-tuning dataset’s sequence clusters to prevent information leakage, yielding \textbf{8731}, \textbf{410}, and \textbf{813} sequences for the Pre-training, RL Fine-tuning, and SOLD TEST datasets, respectively.

Sequence length distributions were analyzed for the Pre-training, RL Fine-tuning, and SOLD TEST datasets, categorizing sequences as short ($\leq 64$ nucleotides), medium ($64 < x \leq 128$ nucleotides), and long ($128 < x \leq 512$ nucleotides). Sequences exceeding 512 nucleotides or containing non-standard nucleotides (non-A, U, C, G) were filtered out, resulting in \textbf{7067}, \textbf{389}, and \textbf{766} sequences for the Pre-training, RL Fine-tuning, and SOLD TEST datasets, respectively. The CASP15 TEST dataset underwent similar filtering for sequences longer than 512 nucleotides and non-AUCG nucleotides, serving as the external test dataset. This process, encompassing clustering, deduplication, and sequence filtering, ensured the dataset’s diversity and consistency, providing a robust foundation for the LDM pre-training, RL fine-tuning, and evaluation.

\begin{figure}[h]
\centering
\includegraphics[width=0.4\textwidth]{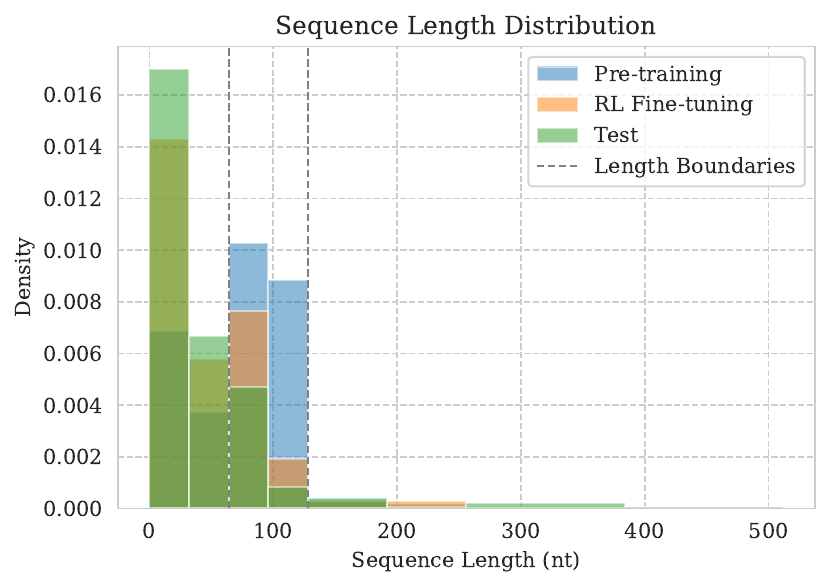}
\caption{Sequence length distribution across Pre-training, RL Fine-tuning, and SOLD TEST datasets.}
\label{fig:seq_len_dist}
\end{figure}

\section*{B. Evaluation Metrics}

This section outlines the evaluation metrics used to assess the performance of the SOLD algorithm in RNA design, categorizing them into three groups: one-dimensional metrics for nucleotide-level accuracy, two-dimensional metrics for secondary structure evaluation, and three-dimensional metrics for tertiary structure fidelity.

\subsection*{B.1 One-Dimensional Metrics}

The one-dimensional metrics focus on the nucleotide-level accuracy of generated RNA sequences, providing insights into sequence prediction quality. The Sequence Recovery (\textbf{Sequence Recovery}) metric quantifies the nucleotide-level prediction accuracy for a single RNA sequence, defined as the proportion of correctly predicted nucleotides:

\begin{equation}
\text{Sequence Recovery} = \frac{1}{N} \sum_{i=1}^{N} \mathbf{1} \left[ \mathcal{S}_i = \arg\max \mathcal{P}(\mathcal{S}_i) \right]
\end{equation}

where \(N\) is the sequence length, \(\mathcal{S}_i\) is the ground truth nucleotide at position \(i\), \(\mathcal{P}(\mathcal{S}_i)\) is the predicted probability distribution over nucleotides (A, U, C, G), and \(1[\cdot]\) is the indicator function (1 if predicted matches ground truth, 0 otherwise). This metric is computed per sequence during validation and averaged across the dataset to assess overall per-sequence accuracy, reflecting the model's ability to reproduce individual RNA sequences. 

The Nucleotide Recovery (\textbf{NT Recovery}) metric measures the overall nucleotide-level prediction accuracy across the entire dataset, defined as:
\begin{equation}
\text{NT Recovery} = \frac{\sum_{\text{seq}} \sum_{i=1}^{N_s} \ 1[\mathcal{S}_{s,i} == \arg \max \mathcal{P}(\mathcal{S}_{s,i})]}{\sum_{\text{seq}} N_s},
\end{equation}
where \(N_s\) is the length of sequence \(s\), \(\mathcal{S}_{s,i}\) is the ground truth nucleotide at position \(i\) in sequence \(s\), and \(\mathcal{P}(\mathcal{S}_{s,i})\) is the predicted probability distribution. This metric aggregates accuracy across all sequences, complementing Sequence Recovery by providing a dataset-wide assessment of nucleotide prediction performance.

\subsection*{B.2 Two-Dimensional Metrics}

The two-dimensional metrics evaluate the secondary structure of generated RNA sequences, assessing functional similarity and thermodynamic stability. The Secondary Structure Similarity (\textbf{SS}) metric measures the similarity between the predicted secondary structure of the generated sequence and the ground truth, using the ViennaRNA~\cite{lorenz2011} package to predict structures in dot-bracket notation (dots for unpaired nucleotides, brackets for paired ones). It is calculated as:
\begin{equation}
\text{SS} = \frac{\text{matching}}{\text{length}} \times 100,
\end{equation}
where \(\text{matching}\) is the number of positions where predicted and ground truth structures agree (both paired or unpaired), and \(\text{length}\) is the sequence length. Expressed as a percentage, SS reflects the functional similarity of the generated sequence's secondary structure to the target. 

The Minimum Free Energy (\textbf{MFE}) metric assesses the thermodynamic stability of the generated sequence's secondary structure, also computed using ViennaRNA. It returns the minimum free energy (in kcal/mol), with lower values indicating greater stability, complementing SS by providing an energetic perspective crucial for RNA functionality.

\subsection*{B.3 Three-Dimensional Metrics}

The three-dimensional metrics evaluate the tertiary structure fidelity of generated RNA sequences, comparing predicted 3D structures to ground truth using RhoFold~\cite{shen2024accurate}. The Root Mean Square Deviation (\textbf{RMSD}) metric quantifies the average deviation between predicted and ground truth 3D structures, focusing on C4' backbone atoms. After alignment via the Kabsch algorithm~\cite{kabsch1976solution}, it is defined as:
\begin{equation}
\text{RMSD} = \sqrt{\frac{1}{N} \sum_{i=1}^N \|\text{true\_atom}_i - \text{pred\_atom}_i\|^2 + \epsilon},
\end{equation}
where \(N\) is the number of aligned C4' atoms, \(\text{true\_atom}_i\) and \(\text{pred\_atom}_i\) are the coordinates of the \(i\)-th C4' atom, and \(\epsilon = 10^{-6}\) prevents numerical instability. A lower RMSD indicates better structural alignment.

The Local Distance Difference Test (\textbf{LDDT}) metric assesses local structural similarity based on C4' atoms, comparing pairwise distances within a 15.0 Å cutoff. It aggregates the fraction of distances deviating by less than 0.5, 1.0, 2.0, and 4.0 Å (each weighted 0.25), averaged over valid pairs. Scores range from 0 to 1, with higher values indicating better local geometry preservation, complementing RMSD's global focus.

\section*{C. LDM Details}


To provide a comprehensive understanding of the Latent Diffusion Model (LDM), this appendix details its network architecture, abalation studies, and comparison with alternative approaches. The LDM generates RNA sequences in a variable-length latent space, leveraging RNA-FM embeddings and backbone geometric features to ensure structural consistency.

\subsection*{C.1 LLM-Related RNA Algorithms}

Large Language Models (LLMs) enhance RNA analysis with large RNA sequence datasets (e.g., 36 million for RiNaLMo~\cite{rafael2024}), for example, RNA-FM~\cite{chen2022interpretable} allows structure and function predictions from unannotated data, CodonBERT~\cite{Li2023} optimizes mRNAs, RNA-GPT~\cite{xiao2024} generates multi-modal alignment sequences, and Nucleotide Transformer~\cite{Hugo2025} and Evo~\cite{Nguyen2024} extend LLM to DNA/RNA applications. However, the 3D design of the RNA lags behind the AlphaFold2~\cite{jumper2021} focused on proteins due to data scarcity.

\subsection*{C.2 Network Architecture}

\begin{algorithm}
\caption{LDM Algorithm}
\label{alg:ldm}
\begin{algorithmic}[1]
\STATE \textbf{Input}: Target distribution \(p(h, c)\) on RNA-FM target embedding \( h \in \mathbf{R}^{L \times 640} \) and backbone \( c \), timestep \( T \), noise schedule \( \beta_t \), pre-trained encoder \( \text{Enc}_\psi \), and decoder \( \text{Dec}_\phi \)
\STATE Initialize MLP encoder denoising network \( \pi_\theta \)
\WHILE{Training}
    \STATE Sample data with backbone conditioning from the dataset: \((t, h, c)\sim Uniform({1, ..., T}), p(h, c)\)
    \STATE Compressed embeddings: \( z_0 = \text{Enc}_\psi(h) \)
    \STATE Generate noisy latent: \( x_t \sim q(x_t \mid z_0)\)
    \STATE Predict target embedding: \( \hat{z}_0 = \pi_\theta(x_t, t, c) \)
    \STATE Sample reverse process: \( z_{t-1}\) using Equation~\ref{eq:reverse_diffusion}
    \STATE Compute loss: \( L = \| z_0 - \hat{z}_0 \|^2 -\sum_{i=1}^L \log p_\phi(s_i \mid \text{Dec}_\phi(\hat{z}_0)_i) \)
    \STATE Update parameters: \(\theta \gets \theta - \eta \nabla_\theta L\)
\ENDWHILE
\STATE \textbf{return} \( \pi_\theta \)
\end{algorithmic}
\end{algorithm}

The Latent Diffusion Model (LDM) generates RNA sequences aligned with backbone structures, leveraging RNA-FM embeddings and geometric features. 

Figure~\ref{fig:LDM} illustrates the architecture, inspired by RiboDiffusion~\cite{huang2024ribodiffusion}, comprising an MLP encoder, a diffusion network, and an MLP decoder. Unlike RiboDiffusion's one-hot sequence input $(L, 4)$, SOLD utilizes RNA-FM embeddings $(L, 640)$, compressed to a latent space $(L, 32)$ via a 3-layer MLP encoder. The diffusion network consists of a 4-layer GVP-GNN, incorporating geometric vector perceptrons~\cite{jing2021learning} (node hidden dimension 512, edge hidden dimension 128, including dihedral angle features) and a 8-layer Transformer (embedding dimension 512, 16 attention heads, dropout 0.2). The GVP-GNN processes backbone geometric features (e.g., bond angles, bond lengths), while the Transformer captures sequence dependencies, predicting the clean latent embedding. A 3-layer MLP decoder reconstructs the final denoised latent embedding into sequence probabilities $(L, 4)$, ensuring length consistency with the input structure. Algorithm~\ref{alg:ldm} presents the LDM training process.

\subsection*{C.3 Abalation Study on Latent Embedding Dimension}

\begin{table}[ht]
  \centering
  \begin{tabular}{l c}
    \toprule
    \textbf{Latent Dimension ($D$)} & \textbf{Sequence Recovery (SOLD TEST)} \\
    \midrule
    8  & 0.5161 \\
    16 & 0.8746 \\
    32 & 0.9768 \\
    64 & 0.9955 \\
    128 & 0.9989 \\
    \bottomrule
  \end{tabular}
  \caption{Test set recovery rates for different latent dimensions $D \in \{8, 16, 32, 64, 128\}$.}
  \label{tab:ablation_dimension}
\end{table}

\begin{figure}[h]
  \centering
  \includegraphics[width=0.4\textwidth]{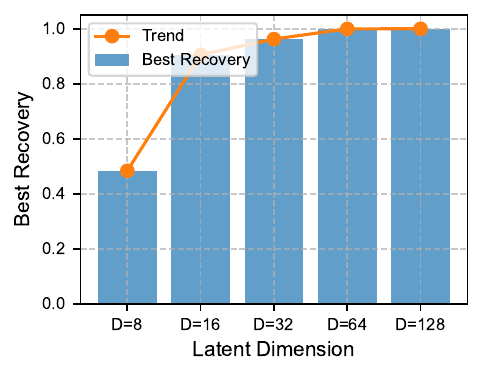}
  \caption{Best recovery rate across different latent dimensions $D \in \{8, 16, 32, 64, 128\}$.}
  \label{fig:recovery_vs_dimension}
\end{figure}

To determine an appropriate latent dimension for subsequent diffusion experiments, we evaluated the impact of compressing RNA-FM embeddings from $(L, 640)$ to $(L, D)$ with $D \in \{8, 16, 32, 64, 128\}$. We measured the best recovery rate on the validation set, defined as the highest recovery rate achieved during training, across different latent dimensions. Figure~\ref{fig:recovery_vs_dimension} illustrates the trend of best recovery rates on the validation set as a bar-line chart, showing a general increase with larger dimensions. On the \textbf{SOLD TEST} dataset, Table~\ref{tab:ablation_dimension} provides detailed recovery rates. The recovery rate improves significantly from $D=8$ to $D=64$, but increasing the dimension to $D=128$ yields only a marginal gain. Considering computational efficiency, $D=32$ and $D=64$ strike a strong balance, offering effective latent representations for diffusion-based generation with manageable training times, whereas $D=8$ and $D=16$ suffer from notable information loss, and $D=128$ significantly increases training time for minimal performance improvement.

\subsection*{C.4 Optimal Latent Dimension for Generation}

\begin{figure}[h]
    \centering
    \includegraphics[width=0.4\textwidth]{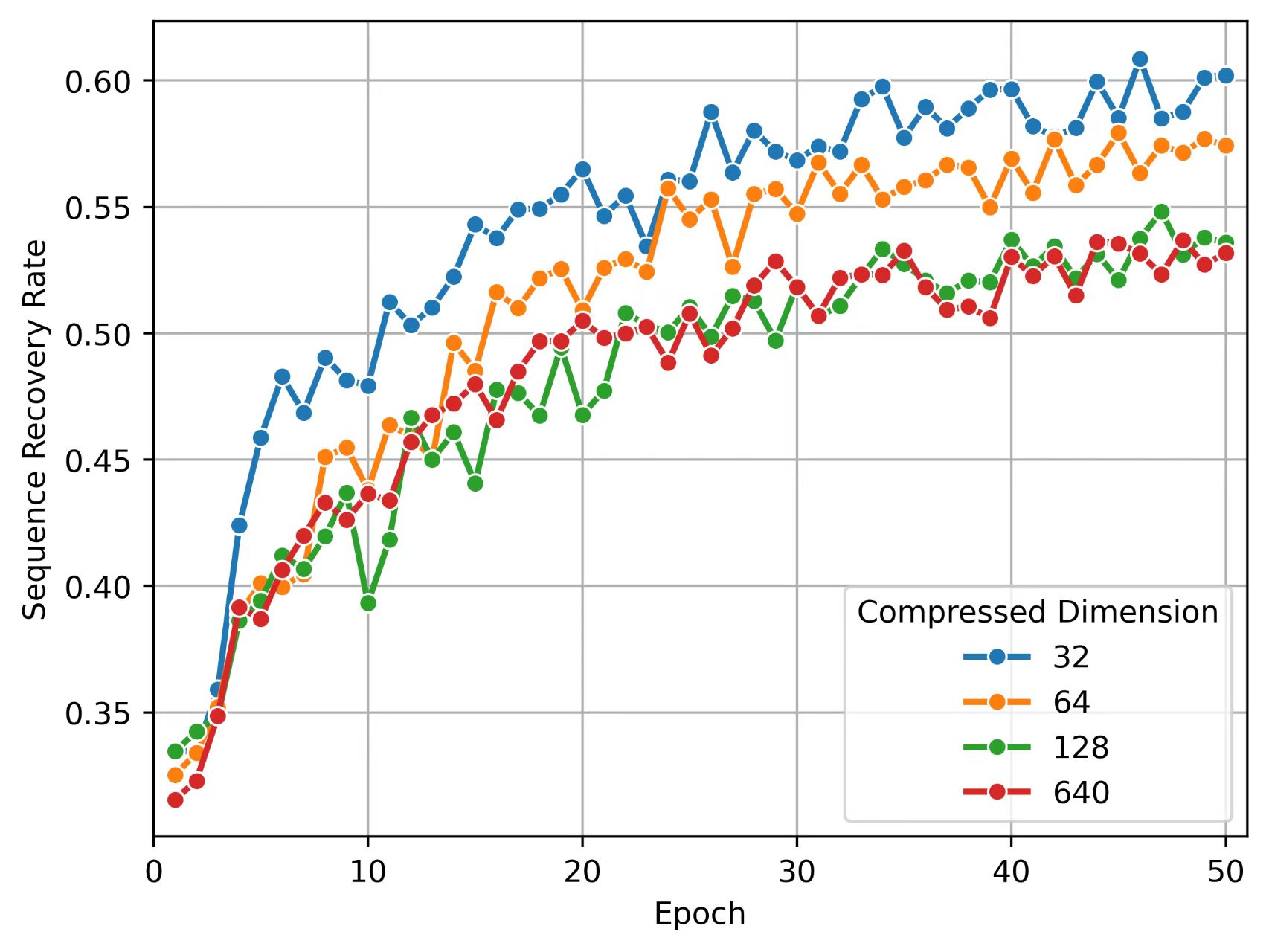}
    \caption{Generative performance of LDM across different latent dimensions $D$.}
    \label{fig:gen-dim}
\end{figure}

To assess the impact of latent dimension \( D \) on generative performance, we extended the ablation study to \( D \in \{32, 64, 128, 640\} \), evaluating sequence recovery (Sequence Recovery) on generated RNA sequences. Figure~\ref{fig:gen-dim} plots the performance trends on the test set, revealing that generative quality peaks at \( D = 32 \). Increasing \( D \) beyond 32 leads to a decline, suggesting that a larger latent space introduces noise and complexity, hindering generation. This mirrors the optimization dilemma noted by Yao et al.~\cite{yao2025vavae} in image generation, where higher dimensions improve reconstruction but degrade generation due to unconstrained representations. For LDM, \( D = 32 \) balances generative quality and training efficiency effectively.

\subsection*{C.5 Training Details}

The LDM was trained using a standard latent diffusion pipeline, training for 50 epochs with a batch size of 64, utilizing 8 parallel workers and the AdamW optimizer with a learning rate of 0.0001 and weight decay of 0.01. The diffusion process operated over 100 steps. Early stopping was applied with a patience of 10 epochs, monitoring validation recovery with a minimum boost of 0.005. The LDM employed a cosine beta schedule to control noise addition, while sampling procedure supported DDPM and DDIM methods, with DDIM using an eta of 1.0 for stochasticity. 

\section*{D. SOLD Details}

This section provides a theoretical foundation for the SOLD algorithm by deriving an approximate variational objective for its reward maximization, focusing on long-term rewards at $t=0$ and incorporating instantaneous short-term rewards at $t-1$ via a piecewise reward switching mechanism. We integrate the DDIM~\cite{song2020denoising} framework's one-step sampling for long-term rewards and the DDPM~\cite{ho2020denoising} framework's one-step sampling for short-term rewards to construct this objective, validated by ablation studies on SOLD TEST dataset demonstrating improved metrics and faster convergence.

\subsection*{D.1 Approximate Variational Objective for Reward Optimization}

This part derives an approximate variational objective for the SOLD algorithm's reward maximization, ensuring practical convergence for its fine-tuning phase. We leverage the DDIM framework's one-step sampling from $z_t$ to $z_0'$ for long-term rewards and the DDPM framework's one-step sampling from $z_t$ to $z_{t-1}$ for short-term rewards, both designed for efficiency rather than the full multi-step ELBO used in methods like DDPO~\cite{black2023training} and DPOK~\cite{Fan2023DPOK}. Rewards are integrated using a piecewise function, prioritizing short-term rewards in early denoising steps and long-term rewards in later steps.

The SOLD objective is defined as:
\begin{equation}
\begin{aligned}
\mathcal{J}_{\text{SOLD}}(\theta) = \mathbf{E}_{t \sim \mathcal{U}[1, T], c, z_t, z_{t-1}', z_0' \sim p_\theta(z_{t-1}', z_0' \mid z_t, c)} \big[ r_{\text{total}}(t) \big],
\end{aligned}
\end{equation}
where:
\begin{itemize}
    \item $t$ is a uniformly sampled timestep,
    \item $c$ is the conditioning input (RNA structure),
    \item $z_t$ is the noised sequence,
    \item $z_{t-1}' \sim p_\theta(z_{t-1}' \mid x_t, c)$ is the optimized intermediate sequence.
    \item $z_0' \sim p_\theta(x_0' \mid z_t, c)$ is the final decoded sequence.
    \item $r_{\text{total}}(t)$ is a piecewise reward function, defined as:
    \begin{equation}
    \small
    r_{\text{total}}(t) = \left\{
    \begin{array}{@{}ll@{}}
    R_i(z_{t-1}') & t \geq \tau, \\
    R_i(z_0') & t < \tau,
    \end{array}
    \right.
    \end{equation}
where $\tau$ is a tunable threshold parameter that determines the switching point between short-term and long-term rewards, $R_i$ represents the reward function for objective $i \in \{\text{MFE}, \text{SS}, \text{LDDT}\}$, and both $z_{t-1}'$ and $z_0'$ are derived from the network's predictions conditioned on $z_t$ and $c$. Specifically, $R_i(z_{t-1}')$ denotes the short-term reward, evaluated on the intermediate sequence $z_{t-1}'$ and typically utilized in the early denoising steps to guide the process toward feasible intermediate states, while $R_i(z_0')$ represents the long-term reward, assessed on the final decoded sequence $z_0'$ and employed in the later denoising steps to optimize the overall stability and quality of the generated sequence.
\end{itemize}

The forward diffusion process follows:
\begin{equation}
q(z_t \mid z_{t-1}) = \mathcal{N}(z_t; \sqrt{1 - \beta_t} z_{t-1}, \beta_t I),
\end{equation}
with the cumulative noised sequence:
\begin{equation}
z_t = \sqrt{\bar{\alpha}_t} z_0 + \sqrt{1 - \bar{\alpha}_t} \epsilon, \quad \epsilon \sim \mathcal{N}(0, I),
\end{equation}
where $\bar{\alpha}_t = \prod_{s=1}^t (1 - \beta_s)$ aggregates the variance schedule $\beta_t$.

The reverse sampling processes are approximated as:
\begin{itemize}
    \item short-term (from $t$ to $t-1$): $z_{t-1}' = \sqrt{\bar{\alpha}_{t-1}} \hat{z}_\theta(z_t, t) + \sqrt{\beta_t} \epsilon$,
    \item long-term (from $t$ to $0$): $z_0' = \sqrt{\bar{\alpha}_0} \hat{z}_\theta(z_t, t) + \sigma_{t,0} \epsilon$,
\end{itemize}
where $\bar{\alpha}_0 = 1$, $\hat{z}_\theta(z_t, t)$ is the network's prediction of the clean sequence, $\beta_t$ is the noise schedule, and $\sigma_{t,0}$ is derived from the variance schedule.

The reverse distributions are modeled as:
\begin{equation}
\small
\begin{array}{ll}
p_\theta(z_{t-1}' \mid z_t) = \mathcal{N}(\hat{z}_\theta(z_t, t), \beta_t I), \\
p_\theta(z_0' \mid z_t) = \mathcal{N}(\hat{z}_\theta(z_t, t), \sigma_{t,0}^2 I),
\end{array}
\end{equation}
where $\beta_t$ and $\sigma_{t,0}^2$ reflect the respective noise levels.

The approximate variational objective is based on the reward-weighted log-likelihood:
\begin{equation}
J_{SOLD} = \mathbf{E}_{t, c, z_t, z_{t-1}', z_0'} \left[ r_{\text{total}}(t) \log p_\theta(z_s') \right],
\end{equation}
where $z_s'$ is $z_{t-1}'$ for $t \geq \tau$ and $z_0'$ for $t < \tau$.

The gradient of the objective is:
\begin{equation}
\nabla_\theta J_{SOLD} \propto \mathbf{E}_{t, c, z_t, z_{t-1}', z_0'} \left[ r_{\text{total}}(t) \nabla_\theta \log p_\theta(z_s' \mid z_t) \right].
\end{equation}
Under the condition where $\tau = T$ (focusing solely on long-term rewards), $r_{\text{total}}(t) = R_i(z_0')$ for all $t$, reducing the SOLD objective to:
\begin{equation}
\nabla_\theta J_{SOLD} \approx \mathbf{E}_{t, c, z_t, z_0'} \left[ R_i(z_0') \nabla_\theta \log p_\theta(z_0' \mid z_t) \right].
\end{equation}
The evidence lower bound (ELBO) for $\log p_\theta(z_0')$ is:
\begin{equation}
\log p_\theta(z_0') \geq \mathbf{E}_{q(z_{1:T} \mid z_0)} \left[ \log \frac{p_\theta(z_{0:T}' \mid c)}{q(z_{1:T} \mid x_0)} \right].
\end{equation}
Given DDIM's one-step sampling efficiency, we use a single-step approximation:
\begin{equation}
\log p_\theta(z_0') \geq \mathbf{E}_{q(z_t \mid z_0)} \left[ \log \frac{p_\theta(z_0' \mid z_t, c) p(z_t)}{q(z_t \mid z_0)} \right].
\end{equation}
Simplifying, we obtain:
\begin{equation}
\log p_\theta(z_0') \geq \mathbf{E}_{q(z_t \mid z_0)} \left[ \log p_\theta(z_0' \mid z_t, c) \right] + C,
\end{equation}
where $C$ is a $\theta$-independent constant:
\begin{equation}
C = \mathbf{E}_{q(z_t \mid z_0)} \left[ \log p(z_t) \right] - \mathbf{E}_{q(z_t \mid z_0)} \left[ \log q(z_t \mid z_0) \right].
\end{equation}
The gradient of the DDPO objective, based on the DDPM full-step sampling to obtain $z_0^{\text{full}}$, is:
\begin{equation}
\nabla_\theta J_{\text{DDPO}} \propto \mathbf{E}_{q(z_t \mid z_0)} \left[ R_i(z_0^{\text{full}}) \nabla_\theta \log p_\theta(z_0^{\text{full}} \mid z_t) \right].
\end{equation}
For SOLD, the long-term reward is derived from the DDIM one-step sampling, yielding $z_0'$. Under the assumption that the DDPM full-step sampling converges to a distribution close to the DDIM one-step sampling for $z_0'$, with bounded prediction error, we approximate $z_0^{\text{full}} \approx z_0'$. Under this approximation, we obtain:
\begin{equation}
\nabla_\theta J_{\text{DDPO}} \approx \nabla_\theta J_{\text{SOLD}}(\theta).
\end{equation}
This demonstrates that, in the extreme case where $\tau = T$ and DDPM's full-step sampling closely approximates DDIM's one-step sampling for the final sequence $z_0'$, SOLD's optimization reduces to an equivalent form of DDPO's objective, leveraging the efficiency of a single-step process to reduce complexity per iteration. Moreover, experimental results reveal that integrating short-term rewards with long-term rewards through the tunable threshold $\tau$ yields superior performance. Algorithm~\ref{alg:sold} illustrates the single-step optimization process in SOLD.

\begin{algorithm}[t!]
\caption{SOLD Algorithm}
\label{alg:sold}
\begin{algorithmic}[1]
\STATE \textbf{Input}: Target distribution \(p(h, c)\) on compressed target embedding \( z \in \mathbf{R}^{L \times D} \) and backbone \( c \), pre-trained model \(\pi_\theta\), timestep \( T \), reward function \( R_i \), weighting functions \( w(t) \) and \( u(t) \), learning rate \( \eta \), KL weights \(\lambda_{\text{ref}}\)
\STATE Initialize \(\theta_{\text{old}} \gets \theta\)
\WHILE{\(\theta\) not converged}
    \STATE Sample data with backbone conditioning from the dataset: \((t, z_0, c)\sim Uniform({1, ..., T}), p(z, c)\)
    \STATE Calculate noisy latent \( z_t \sim q(z_t \mid z_0) \)
    \STATE Predict target embedding \(\hat{z}_0 = \pi_\theta(z_t, t, c)\)
    \STATE Calculate optimized embedding \( z_0'\) using Equation (\ref{eq:ddim}), decode \( s_0' \) from \( z_0' \) to calculate long-term reward
    \STATE Calculate \( z_{t-1} \) from \(\hat{z}_0\) as the reverse process defined in Equation (\ref{eq:reverse_diffusion}), then decode \( z_{t-1} \) into \( s_{t-1} \) to calculate short-term reward
    \STATE Calculate reward \( r_{\text{total}}(t) \) using Equation (\ref{eq:rewards}), and then normalized rewards across batch samples to reduce high variance.
    \STATE Calculate importance weight \( w = \frac{p_\theta(z_0' \mid z_t, c, t)}{p_{\text{ref}}(z_0' \mid z_t, c, t)} \)
    \STATE Calculate gradient:\\
    \setlength\leftskip{2em}{\( g = w \cdot \nabla_\theta \log p_\theta(z_0' \mid z_t, c, t) \cdot r_{\text{total}}(t) \)}
    \STATE \setlength\leftskip{0em}{Calculate KL penalties:}\\
    \setlength\leftskip{2em}{\(\text{KL}_{\text{ref}} = D_{\text{KL}}(p_\theta| p_{\text{ref}})\)}
    \STATE \setlength\leftskip{0em}{Update parameters:} \\
    \setlength\leftskip{2em}{\(\theta \gets \theta + \eta \left( g - \lambda_{\text{ref}} \nabla_\theta \text{KL}_{\text{ref}} \right)\)}
    \STATE \setlength\leftskip{0em}{Update old policy: \(\theta_{\text{old}} \gets \theta\)}
\ENDWHILE
\STATE \textbf{return}: \(\pi_\theta\)
\end{algorithmic}
\end{algorithm}

\subsection*{D.2 Piecewise Reward Strategy Abalation}

To enhance the optimization of the SOLD algorithm, we introduce a piecewise reward strategy that integrates instantaneous rewards \( r_t \) (via DDPM one-step sampling) and long-term rewards \( r_0 \) (based on the final decoded sequence \( x_0' \) via DDIM one-step sampling). This approach utilizes the dense feedback of \( r_t \) in early denoising steps for effective initial guidance and the stability of \( r_0 \) in later steps for final optimization, with the transition controlled by a tunable threshold parameter \(\tau\).

We conducted an ablation study to evaluate the performance of different reward configurations across three objectives: MFE, SS, and LDDT. Each objective was tested with five settings: long-term reward only, short-term reward only, and mixed reward strategies with \(\tau = 90\), \(\tau = 60\), and \(\tau = 20\), where the total diffusion steps are 100. The mixed reward strategy prioritizes short-term rewards for \( t \geq \tau \) and transitions to long-term rewards for \( t < \tau \). 

Table~\ref{tab:reward_ablation} summarizes the experimental results for each setting on the SOLD TEST dataset, revealing notable trends across the objectives. For MFE and LDDT, the mixed strategy with a higher \(\tau\) value tends to outperform single-reward settings, suggesting that a later transition to long-term rewards enhances stability. For SS, an intermediate \(\tau\) value appears to strike the best balance between short-term guidance and long-term optimization, indicating a more effective integration of both reward types.These observations suggest that the mixed reward approach adapts effectively to each objective, with the optimal \(\tau\) varying based on the specific requirements of stability, structure, and accuracy.

\begin{table}[h]
    \centering
    \begin{tabular}{lc}
        \toprule
        Setting         & SOLD TEST \\
        \midrule
        \multicolumn{2}{c}{\textbf{MFE Training (MFE reward) \(\downarrow\)}} \\
        \midrule
        Long-term only  & -17.2421 \\
        Short-term only & -17.7265 \\
        \(\tau = 90\)   & \textbf{-19.7428} \\
        \(\tau = 60\)   & -18.7498 \\
        \(\tau = 20\)   & -17.4660 \\
        \midrule
        \multicolumn{2}{c}{\textbf{SS Training (SS reward) \(\uparrow\)}} \\
        \hline
        Long-term only  & 0.7494 \\
        Short-term only & 0.7522 \\
        \(\tau = 90\)   & 0.7496 \\
        \(\tau = 60\)   & \textbf{0.7551} \\
        \(\tau = 20\)   & 0.7548 \\
        \midrule
        \multicolumn{2}{c}{\textbf{LDDT Training (LDDT reward) \(\uparrow\)}} \\
        \midrule
        Long-term only  & 0.6326 \\
        Short-term only & 0.6313 \\
        \(\tau = 90\)   & \textbf{0.6384} \\
        \(\tau = 60\)   & 0.6371 \\
        \(\tau = 20\)   & 0.6338 \\
        \bottomrule
    \end{tabular}
    \caption{Ablation study results for piecewise reward strategies across MFE, SS, and LDDT objectives with \(\tau = 90\), \(\tau = 60\), and \(\tau = 20\) on the SOLD TEST dataset.}
    \label{tab:reward_ablation}
\end{table}

\begin{figure*}[t]
    \vspace*{-10pt} 
    \centering
    \includegraphics[width=\textwidth]{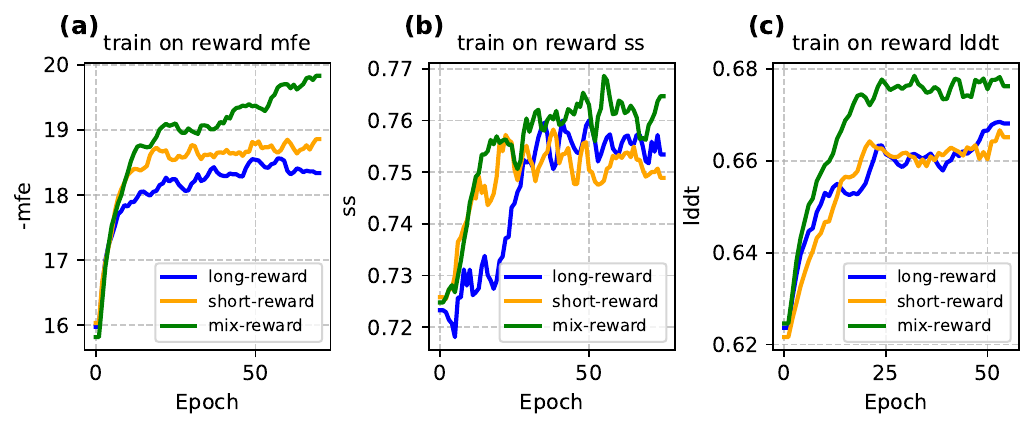}
    \caption{Validation results for \textbf{long-term reward}, \textbf{short-term reward}, and \textbf{mixed reward} settings with MFE, SS, and LDDT as reward objectives: (a) MFE, (b) SS, (c) LDDT.}
    \label{fig:reward_ablation}
\end{figure*}

Figure~\ref{fig:reward_ablation} illustrates the reward trends for MFE, SS, and LDDT over epochs. These visualizations highlight the complementary dynamics of the mixed reward approach, bridging the early exploration of short-term rewards and the later exploitation of long-term rewards. with hyperparameters tuned as follows: for MFE and LDDT, long-term reward dominates the last 90 steps and short-term the first 10 steps; for SS, long-term dominates the last 60 steps and short-term the first 40 steps.

\subsection*{D.3 Training Details}

For training the SOLD algorithm, we implemented an online reinforcement learning (RL) training pipeline using the Proximal Policy Optimization (PPO) algorithm, incorporating techniques inspired by DPOK to ensure stability and efficiency. The training process spanned 100 epochs, with a batch size of 32 sequences, leveraging 8 parallel workers for data loading. Time steps \( t \) were sampled uniformly from \([1, T]\), where \( T = 100 \), as specified in the configuration.

The reward computation followed a piecewise reward strategy, utilizing a threshold-based approach where short-term rewards dominate for \( t > \tau \) and long-term rewards for \( t \leq \tau \). Reward components were computed as a weighted combination of multiple objectives: \( r = w_{\text{SS}} \cdot \text{SS} + w_{\text{MFE}} \cdot \text{MFE} + w_{\text{LDDT}} \cdot \text{LDDT} \), with weights \( w_{\text{SS}} = 0.0 \), \( w_{\text{MFE}} = 1.0 \), and \( w_{\text{LDDT}} = 0.0 \) for the specific run, adjusted based on the objective (e.g., MFE focused in the provided configuration). Rewards were calculated using 16 parallel jobs for efficiency.

The PPO optimization employed a clipped surrogate loss with a clip range of 0.0001. Advantages were estimated as \( A = \frac{r - \mu_r}{\sigma_r + 10^{-8}} \), where \( r \) is the reward, and \( \mu_r \) and \( \sigma_r \) are the mean and standard deviation of the rewards within a batch, ensuring normalized advantage estimates. Gradients were accumulated over 32 steps before each parameter update, using the AdamW optimizer with a learning rate of \( 10^{-5} \) and a weight decay of 0.001. Gradient clipping was applied with a maximum norm of 1.0 to prevent exploding gradients. Batch normalization weights were frozen to align with the pre-trained model, enhancing stability.

KL regularization was implemented to prevent overfitting and preserve the pre-trained model's capabilities. The KL divergence between the current model \( p_\theta \) and the reference model \( p_{\text{ref}} \) was computed as \( \text{KL} = 0.5 \cdot \mathbf{E}[(\log p_\theta - \log p_{\text{ref}})^2] \), weighted by \( \alpha = 1.0 \), to encourage exploration while maintaining stability. The total loss was formulated as \( L = L_{\text{policy}} + \alpha \cdot \text{KL}_{\text{ref}} \), balancing policy optimization with regularization.

\begin{table*}[t]
    \centering
    \begin{tabular}{lccccc}
    \toprule
        Method         & Sequence Recovery\(\uparrow\) & MFE\(\downarrow\) & SS\(\uparrow\) & RMSD\(\downarrow\) & LDDT\(\uparrow\) \\
        \midrule
        \multicolumn{6}{c}{\textbf{SOLD TEST (short)}} \\
        \midrule
        RhoDesign      & 0.2655 \(\pm\) 0.0492 & -4.4230 \(\pm\) 1.664 & 0.7043 \(\pm\) 0.0788 & 12.2489 \(\pm\) 3.5662 & 0.5513 \(\pm\) 0.0542 \\
        
        RDesign        & 0.4691 \(\pm\) 0.0720 & -3.6376 \(\pm\) 1.5571 & 0.6636 \(\pm\) 0.1293 & 11.8251 \(\pm\) 4.4616 & 0.5787 \(\pm\) 0.0877 \\
        
        gRNAde         & 0.5187 \(\pm\) 0.0837 & -2.9642 \(\pm\) 1.5130 & 0.6034 \(\pm\) 0.1184 & 13.7774 \(\pm\) 4.6913 & 0.5309 \(\pm\) 0.0751 \\
        
        RiboDiffusion  & 0.5459 \(\pm\) 0.0600 & -6.1953 \(\pm\) 1.3701 & 0.8203 \(\pm\) 0.0831 & 8.1286 \(\pm\) 2.1961 & 0.6642 \(\pm\) 0.0535 \\
        
        DRAKES         & 0.4529 \(\pm\) 0.0563 & -6.9797 \(\pm\) 1.2297 & \textbf{0.8369} \(\pm\) 0.0434 & 7.9829 \(\pm\) 1.9218 & 0.6716 \(\pm\) 0.0572 \\
        
        LDM            & \textbf{0.6035} \(\pm\) 0.0354 & -5.6705 \(\pm\) 1.3069 & 0.7906 \(\pm\) 0.0747 & 7.9229 \(\pm\) 1.7964 & 0.6848 \(\pm\) 0.0443 \\
        
        SOLD           & 0.5915 \(\pm\) 0.0468 & \textbf{-7.0453} \(\pm\) 1.3316 & 0.8251 \(\pm\) 0.0769 & \textbf{7.6320} \(\pm\) 2.1259 & \textbf{0.6984} \(\pm\) 0.0508 \\
        \midrule
        \multicolumn{6}{c}{\textbf{SOLD TEST (medium)}} \\
        \midrule
        RhoDesign      & 0.2984 \(\pm\) 0.0291 & -20.1741 \(\pm\) 3.7921 & 0.4919 \(\pm\) 0.1017 & 23.9709 \(\pm\) 3.6882 & 0.3552 \(\pm\) 0.0436 \\
        
        RDesign        & 0.3745 \(\pm\) 0.0366 & -17.2209 \(\pm\) 3.8661 & 0.4637 \(\pm\) 0.1009 & 24.5116 \(\pm\) 3.7689 & 0.3545 \(\pm\) 0.0484 \\
        gRNAde         & 0.4908 \(\pm\) 0.0414 & -18.2801 \(\pm\) 4.2355 & 0.4405 \(\pm\) 0.0870 & 26.0511 \(\pm\) 3.8313 & 0.3410 \(\pm\) 0.0445 \\
        RiboDiffusion  & 0.4048 \(\pm\) 0.0446 & -26.8418 \(\pm\) 4.2664 & 0.6409 \(\pm\) 0.0865 & 20.2351 \(\pm\) 3.5920 & 0.4636 \(\pm\) 0.0616 \\
        
        DRAKES         & 0.4036 \(\pm\) 0.0335 & -23.6893 \(\pm\) 3.8733 & 0.6634 \(\pm\) 0.0620 & 19.2688 \(\pm\) 3.0542 & 0.4751 \(\pm\) 0.0491 \\
        
        LDM            & 0.4754 \(\pm\) 0.0271 & -22.2533 \(\pm\) 3.2741 & 0.5405 \(\pm\) 0.0941 & 22.5351 \(\pm\) 3.0984 & 0.4018 \(\pm\) 0.0555 \\
        
        SOLD           & \textbf{0.5193} \(\pm\) 0.0247 & \textbf{-28.4233} \(\pm\) 3.0945 & \textbf{0.6676} \(\pm\) 0.0488 & \textbf{19.0019} \(\pm\) 3.1009 & \textbf{0.4835} \(\pm\) 0.0557 \\
        \midrule
        \multicolumn{6}{c}{\textbf{SOLD TEST (long)}} \\
        \midrule
        RhoDesign      & 0.3008 \(\pm\) 0.0219 & -85.9170 \(\pm\) 7.8836 & 0.4186 \(\pm\) 0.0569 & 47.2802 \(\pm\) 4.6981 & 0.3228 \(\pm\) 0.0172 \\
        
        RDesign        & 0.3558 \(\pm\) 0.0205 & -84.2757 \(\pm\) 7.1098 & 0.4139 \(\pm\) 0.0541 & 47.2937 \(\pm\) 4.0661 & 0.3197 \(\pm\) 0.0185 \\
        
        gRNAde         & 0.4686 \(\pm\) 0.0222 & -87.1803 \(\pm\) 8.5181 & 0.3974 \(\pm\) 0.0471 & 49.0031 \(\pm\) 4.9398 & 0.3191 \(\pm\) 0.0200 \\
        
        RiboDiffusion  &  0.4034 \(\pm\) 0.0268 & -98.8963 \(\pm\) 8.0511 & 0.5591 \(\pm\) 0.0550 & 43.3506 \(\pm\) 3.5654 & 0.3488 \(\pm\) 0.0228 \\
        
        DRAKES         & 0.3817 \(\pm\) 0.0186 & -81.3017 \(\pm\) 6.4232 & 0.5437 \(\pm\) 0.0475 & 43.0538 \(\pm\) 2.2279 & 0.3490 \(\pm\) 0.0202 \\
        
        LDM            & 0.4672 \(\pm\) 0.0169 & -87.3374 \(\pm\) 6.4215 & 0.4595 \(\pm\) 0.0572 & 47.3536 \(\pm\) 4.5370 & 0.3303 \(\pm\) 0.0219 \\
        
        SOLD           & \textbf{0.4962} \(\pm\) 0.0138 & \textbf{-111.3642} \(\pm\) 5.4817 & \textbf{0.5633} \(\pm\) 0.0488 & \textbf{42.6735} \(\pm\) 3.0404 & \textbf{0.3524} \(\pm\) 0.0220 \\
        \bottomrule
    \end{tabular}
    \caption{Multi-Objective Performance Comparison across Different RNA Sequence Length}
    \label{tab:multi_objective_consolidated}
\end{table*}

\section*{E. Other RNA Inverse Folding algorithms}

This appendix provides detailed implementations and reproduction protocols for state-of-the-art RNA inverse-folding algorithms compared in this study, ensuring fair and reproducible performance evaluation. We include comprehensive descriptions of training pipelines, architectural configurations, datasets, and evaluation metrics for each method, facilitating transparency and validation.

\subsection*{E.1 RhoDesign Algorithm Reproduction Details}
\label{subsec:rhodesign}
To ensure a fair comparison with SOLD, we reproduced the RhoDesign~\cite{wong2024deep} algorithm, originally provided with inference code only (https://github.com/ml4bio/RhoDesign), based on its described variational autoencoder (VAE) framework. Unlike the original implementation, which leveraged RhoFold to predict 3D structures for 369,499 RNAcentral sequences to augment training data, we adhered to a controlled experimental setup. Specifically, we utilized the pre-training dataset detailed in Appendix B as the training set, without incorporating additional predicted structures, to maintain consistency and fairness across all methods evaluated.

Training was implemented in PyTorch, processing PDB files from the pre-training dataset to extract structural coordinates. We employed the Biotite library to ensure consistent atom counts, focusing on key atoms relevant to RNA backbone representation, though specific atom selections (e.g., C4', C1', N1) were aligned with default settings to match the original methodology. The model was trained for 100 epochs with batch size of 32, learning rate of \(10^{-5}\), and the Adam optimizer on an NVIDIA A100 GPU. Evaluation was conducted on the SOLD TEST dataset and the CASP15 TEST dataset, using the sequence recovery rate as the primary metric, with early stopping applied after 10 epochs without improvement (minimum boost threshold of 0.005 on Sequence Recovery). 

\subsection{E.2 RDesign Algorithm Reproduction Details}
To reproduce the RDesign~\cite{tan2024rdesign} algorithm, we utilized its public available source code (https://github.com/A4Bio/RDesign). The model architecture was maintained consistent with the original implementation, featuring a 3-layer MPNN encoder and decoder. Training was conducted on an NVIDIA A100 GPU, using the Adam optimizer with a learning rate of 0.0003, batch size of 32, and 100 epochs. Mixed precision training with GradScaler (initial scale 4096) and a 0.1 dropout rate were applied, also with early stopping after 10 epochs no boost. 

\subsection*{E.3 gRNAde Algorithm Reproduction Details}

To reproduce the gRNAde~\cite{joshi2025grnade} algorithm,  we utilized its open-source code (https://github.com/chaitjo/geometric-rna-design). The training pipeline was implemented with PyTorch Geometric, processing the SOLD pre-training dataset, where each PDB ID corresponds to a single conformation, unlike the original gRNAde setting that leverages multiple conformations per PDB ID to address structural dynamics. The RNAGraphFeaturizer module constructed a 32-nearest-neighbor graph using a 3-bead coarse-grained representation (P, C4', N1/N9). The model replicated a 4-layer multi-state GVP-GNN encoder (node dimensions (128,16), edge dimensions (64,4), dropout 0.5) and a 4-layer autoregressive GVP-GNN decoder with SE(3)-equivariant updates and Deep Set pooling, trained with cross-entropy loss (label smoothing 0.05). Training ran on an NVIDIA A100 GPU for 100 epochs, using the Adam optimizer with a learning rate of 0.0001, a maximum of 3000 nodes per batch, with a DataLoader using 4 workers and BatchSampler.

Evaluation also utilized the SOLD TEST dataset and CASP15 TEST dataset, focusing on recovery rate, with early stopping after 10 epochs (minimum boost 0.005 on recovery as previous algorithms). The reproduced performance validated gRNAde’s design capabilities, adapted to our single-conformation context, serving as a benchmark for SOLD’s enhancements.

\subsection*{E.4 RiboDiffusion Algorithm Reproduction Details}

To reproduce the RiboDiffusion~\cite{huang2024ribodiffusion} algorithm, for tertiary structure-based RNA inverse folding, we utilized the described methodology and available code (https://github.com/ml4bio/RiboDiffusion), noting that RiboDiffusion did not provide training code, with implementation details inferred from the original paper. The training pipeline utilized the SOLD pretraining dataset. The model replicated a dual-module design: a 4-layer GVP-GNN structure module (node dimensions (512,128), edge dimensions (128,1), dropout 0.1) and an 8-layer Transformer sequence module (512-dimensional embedding, 16 attention heads), trained with a variational diffusion SDE (cosine schedule, beta 0.1 to 20) using weighted mean squared error loss. Training ran on an NVIDIA A100 GPU for 1000 epochs, using the AdamW optimizer (learning rate 0.0001, batch size 1, gradient clipping 100), 8-step gradient accumulation, 0.999 EMA decay for stability, and a DataLoader with 0.1Å Gaussian noise.

\subsection*{E.5 DRAKES Algorithm Reproduction Details}

To reproduce the DRAKES~\cite{wang2024finetuning} algorithm, proposed for fine-tuning discrete diffusion models to optimize  reward functions, we utilized the open-source code (https://github.com/ChenyuWang-Monica/DRAKES) and described methodology. The training pipeline, a two-stage process, used the SOLD pre-training dataset. In the first stage, we trained a masked discrete diffusion model, extracting RNA 3D structure coordinates and sequences, with preprocessing including 10-nearest-neighbor geometric features (16 RBF and 16 positional encodings) and 0.1Å Gaussian noise. The model architecture replicated a 3-layer MPNN encoder and 3-layer decoder (node features 128, edge features 128, hidden dimension 128, dropout 0.1), outputting 5 base probabilities, initialized with Xavier uniform distribution.

Training for the first stage ran for 100 epochs, using the AdamW optimizer (learning rate 0.0001, weight decay 0.0001, batch size 4) with 4 workers. In the second stage, we first pre-trained a reward model on the SOLD dataset to predict RNA stability (MFE via RNAfold), using mean pooling over the last layer of the first-stage model followed by an MLP, trained for 100 epochs with the AdamW optimizer (learning rate 0.0001, weight decay 0.0001, batch size 4, early stopping patience 10). The diffusion model was then fine-tuned during the DRAKES optimization phase using the pre-trained diffusion checkpoint and reward checkpoint, with a masked interpolant diffusion (200 timesteps). Fine-tuning ran for 50 epochs with the AdamW optimizer (learning rate 0.0001, weight decay 0.0001, batch size 8), early stopping (patience 10), KL divergence regularization with a proportion of 0.1, and Gumbel-Softmax with a linear temperature schedule.

\section*{F. Exploration of Length Effects and Computational Efficiency}

This section presents a comprehensive evaluation of SOLD for RNA inverse folding, focusing on its multi-objective performance, stability, and computational efficiency across varying sequence lengths. We analyze SOLD and its base LDM against SOTA RNA inverse folding methods, including RhoDesign, RDesign, gRNAde, RiboDiffusion, and DRAKES, on the SOLD TEST Dataset. Test dataset is partitioned into short ($\leq 64$ nucleotides), medium ($64 < x \leq 128$ nucleotides), and long ($128 < x \leq 512$ nucleotides) sequences as described in appendix A. Additionally, we evaluate SOLD's computational efficiency by measuring sampling time and GPU memory usage for generating sequences across diverse RNA structures.

\subsection{F.1 Multi-Objective Performance Analysis of Sequence Length}
We evaluated SOLD across different sequence length to assess its robustness in designing RNA structures of varying complexity. As shown in Table~\ref{tab:multi_objective_consolidated}, LDM achieves the highest Sequence Recovery across short, medium, and long sequence ranges, outperforming baseline methods such as RiboDiffusion and gRNAde. SOLD, enhanced through RL fine-tuning, further improves all metrics, achieving optimal or near-optimal results across all length categories. Notably, in the long sequence range, SOLD significantly surpasses RDesign and gRNAde in MFE, SS, and LDDT, demonstrating its ability to generate sequences with high structural fidelity and biological functionality.

Among baselines, gRNAde and RDesign perform well on short sequences but exhibit declining performance in medium and long ranges. RiboDiffusion and DRAKES show competitive performance in short and medium sequences but fall short of SOLD in long sequences.  In terms of stability, SOLD exhibits the almost lowest standard deviations across all metrics, surpassing LDM and other baselines, indicating high prediction consistency suitable for precision RNA design.

As sequence length increases, Sequence Recovery, SS, and LDDT tend to decrease, MFE becomes more negative, and RMSD increases, reflecting the challenges of complex RNA folding. SOLD effectively mitigates these challenges, maintaining robust performance and stability. 

\subsection*{F.2 Computational Efficiency}

To evaluate the computational efficiency of SOLD, we measured the sampling time and GPU memory usage required to generate eight RNA sequences for PDB structures with varying sequence lengths. As shown in Table~\ref{tab:sampling_time}, sampling time increases moderately with sequence length, remaining within seconds from short (77 nucleotides) to long (501 nucleotides) sequences. GPU memory usage rises slightly but stays manageable across all lengths. 

\begin{table}[h]
    \centering
    \begin{tabular}{lcccc}
    \hline
        \toprule
        PDB ID         & Length & Time (s) & Memory (MB) \\
        \midrule
        2CKY\_1\_A     & 77     & 37.04    & 7734        \\
        7PKT\_1\_1     & 162    & 42.08    & 7974        \\
        7PUA\_1\_CA    & 501    & 50.99    & 8433        \\
        \bottomrule
    \end{tabular}
    \caption{Sampling time and GPU memory usage for generating eight sequences per PDB structure.}
    \label{tab:sampling_time}
\end{table}

\subsection*{F.3 Significance Testing on SOLD TEST Dataset}

To evaluate differences among all methods used for the RNA inverse folding problem, we conducted significance testing across sequence and structure metrics. Due to non-normal distributions (confirmed by Shapiro-Wilk tests, $p < 0.05$), we employed non-parametric tests.

\begin{figure*}[ht]
    \centering
\includegraphics[width=\textwidth]{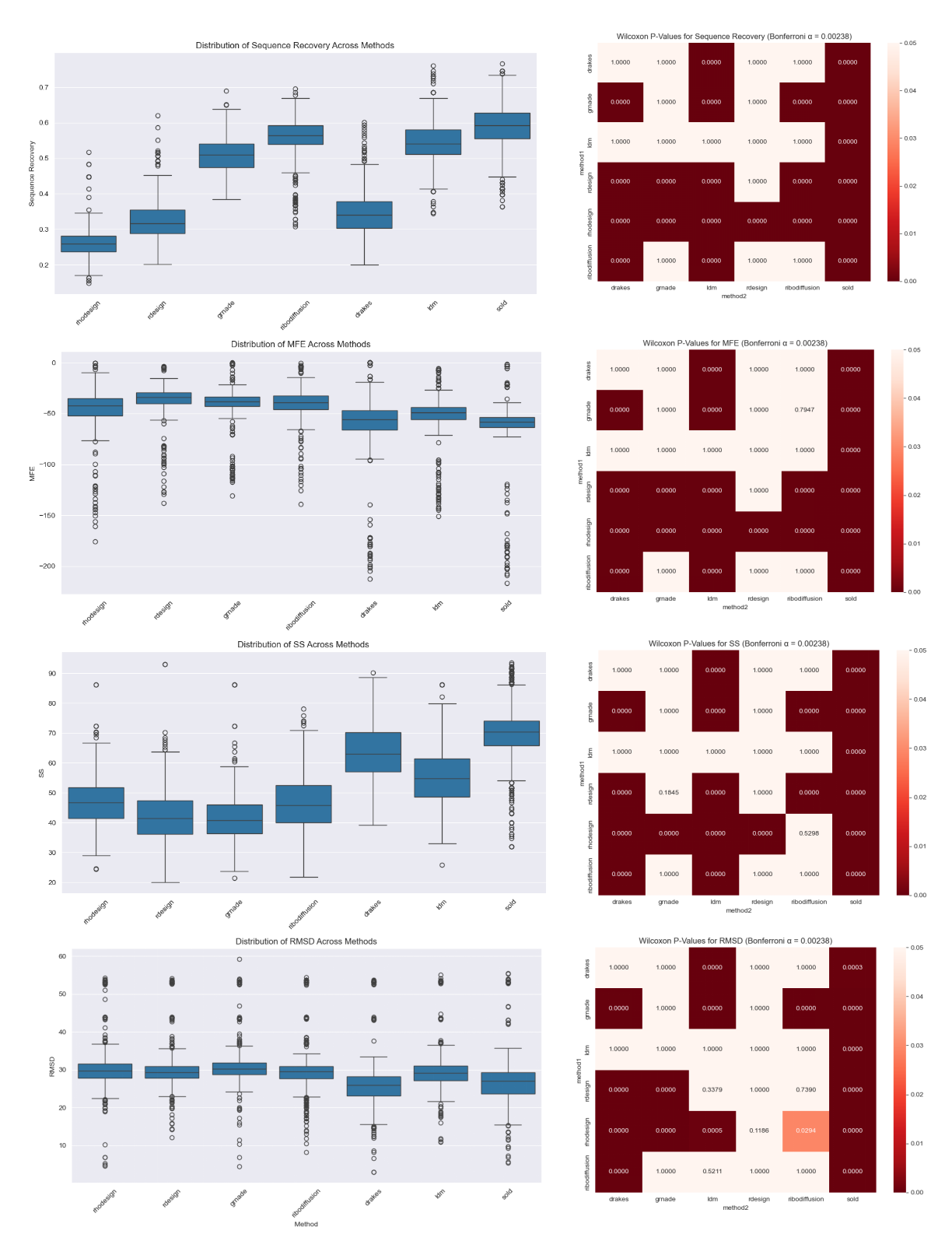}
\caption{Significant for Sequence and Structure Metrics on SOLD TEST dataset}   \label{fig:significant_experiment}
\end{figure*}

The \textbf{Friedman test} was used to assess overall differences across the seven methods for each metric, suitable for paired samples across multiple groups. When significant differences were detected ($p < 0.05$), we performed \textbf{pairwise Wilcoxon signed-rank tests} to compare method pairs (21 comparisons). To account for multiple comparisons, we applied Bonferroni correction ($\alpha = 0.05/21 \approx 0.00238$). Effect sizes were computed as $r = Z / \sqrt{N}$, where $Z$ is the Wilcoxon statistic and $N$ is the number of pairs. Results in Figure~\ref{fig:significant_experiment} are visualized using boxplots for metric distributions and p-value heatmaps for pairwise significance.

The analysis revealed significant differences among methods, with SOLD demonstrating significant superiority over all other methods across all metrics, except for RMSD, where SOLD’s performance was comparable to DRAKES. 

\section*{G. Additional Design Cases}

\begin{figure*}[ht]
    \centering
   \includegraphics[width=\textwidth]{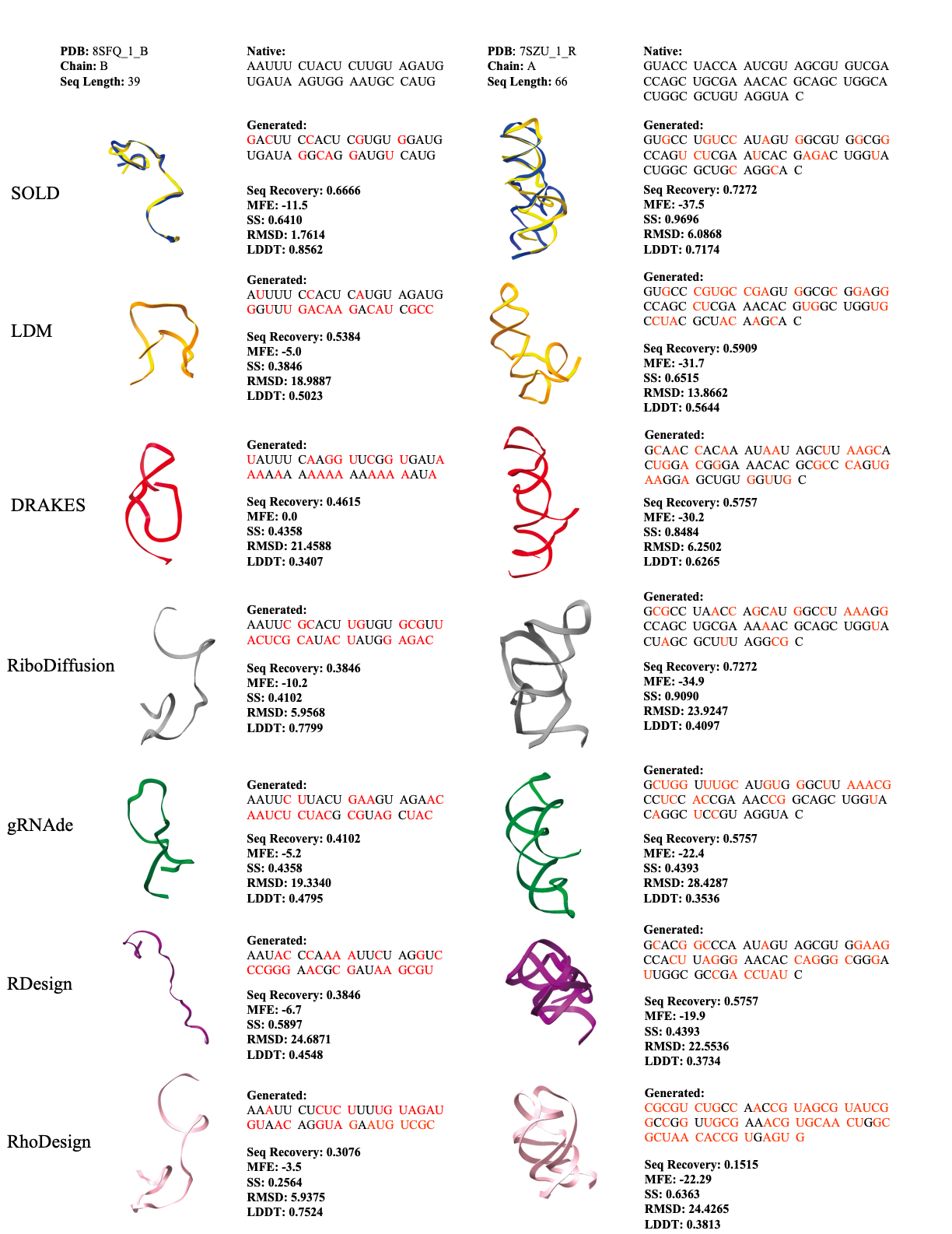}
    \caption{Additional Successful RNA Designs using the SOLD Model (Part I)}, 
  \label{fig:show_case_page1}
\end{figure*}

\begin{figure*}[ht]
    \centering
\includegraphics[width=\textwidth]{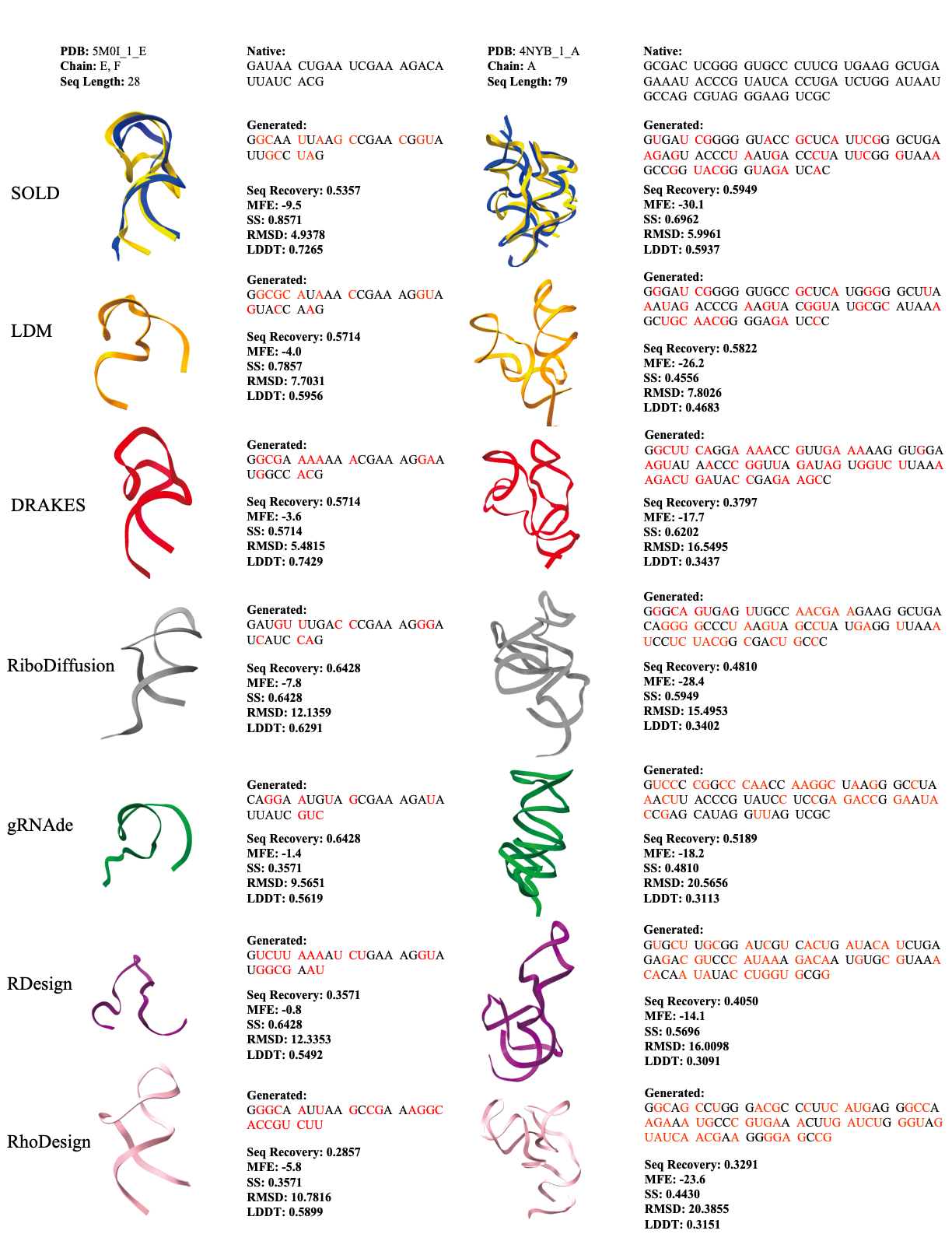}
\caption{Additional Successful RNA Designs using the SOLD Model (Part II)}   \label{fig:show_case_page2}
\end{figure*}

To comprehensively evaluate the performance and robustness of SOLD, this appendix presents additional inverse folding case studies on four distinct RNA targets sourced from the Protein Data Bank (PDB IDs: 5MOJ, 4NYB, 8SFQ, and 7SZU), as shown in Figure~\ref{fig:show_case_page1} and Figure~\ref{fig:show_case_page2}. These targets feature diverse sequence lengths (ranging from 28 to 79 nucleotides) and structural complexities.

For each target, we provide a head-to-head performance comparison of SOLD against six baseline methods: LDM, DRAKES, RiboDiffusion, gRNAde, RDesign and RhoDesign. Because SOLD's predictions are exceptionally accurate, in the visualizations we superimpose its predicted structure (blue) directly onto the ground-truth structure (gold) to highlight this high fidelity. In contrast, the structures generated by baseline methods exhibit significant deviations. This striking difference underscores SOLD's superior accuracy and its reliability in solving the RNA inverse folding problem across a variety of structural challenges.

\end{document}